\documentclass[letterpaper, 10 pt, conference]{ieeeconf}  %

\IEEEoverridecommandlockouts                              %

\usepackage{flushend}

\usepackage{times}
\usepackage{epsfig}
\usepackage{graphicx}
\usepackage{float}
\usepackage{wrapfig}

\usepackage{amsmath,amssymb}

\usepackage{bm,xspace}
\usepackage{comment}
\usepackage{verbatim}
\usepackage{multirow}
\usepackage{multicol}
\usepackage{balance}
\usepackage{url}
\usepackage{booktabs}
\usepackage{etoolbox,siunitx}
\usepackage{calc}
\usepackage{pifont,hologo}
\usepackage{color}

\usepackage[super]{nth}
\usepackage{nicefrac}
\def\tt{\mathbf{t}}

\newcommand{\YesV}{\ding{51}}%
\newcommand{\NoX}{\ding{55}}%

\DeclareMathSymbol{@}{\mathord}{letters}{"3B}

\newcommand\mypara[1]{\vspace{3pt}\noindent\textbf{#1.}}

\def\latex/{\LaTeX}
\def\bibtex/{\hologo{BibTeX}}

\usepackage[official]{eurosym}

\usepackage{subcaption}

\usepackage{hyperref}
\hypersetup{
    colorlinks=true,
    linkcolor=magenta,
    filecolor=magenta,      
    urlcolor=blue,
    pdftitle={OpenBot},
    pdfpagemode=FullScreen,
    }

\newcommand{\ie}{\emph{i.e.\xspace}}
\newcommand{\eg}{\emph{e.g.\xspace}}

\newcommand{\etal}{\emph{et al.\xspace}}

\usepackage[capitalize]{cleveref}
\Crefname{table}{Table}{Tables}
\crefname{table}{Tab.}{Tabs.}
\crefname{section}{Sec.}{Secs.}
\Crefname{section}{Section}{Sections}
\Crefname{figure}{Figure}{Figures}
\crefname{figure}{Fig.}{Figs.}
\crefname{algorithm}{Alg.}{Algs.}
\Crefname{algorithm}{Algorithm}{Algorithms}
\crefname{equation}{Eq.}{Eqs.}
\Crefname{equation}{Equation}{Equations}

\newcommand{\systemName}{OpenBot-Fleet\xspace}
\newcommand{\supp}{appendix\xspace}

\title{\LARGE \bf \systemName: A System for Collective Learning with Real Robots}

\author{Matthias M\"uller, Samarth Brahmbhatt, Ankur Deka, Quentin Leboutet, David Hafner, and Vladlen Koltun \\[8pt]
Intel Labs \\[8pt]
}

\begin{document}

\twocolumn[{%
\renewcommand\twocolumn[1][]{#1}%
\maketitle
 \begin{center}
    \centering
        \captionsetup{type=figure}
        \vspace{-2em}
 	\includegraphics[width=1\textwidth]{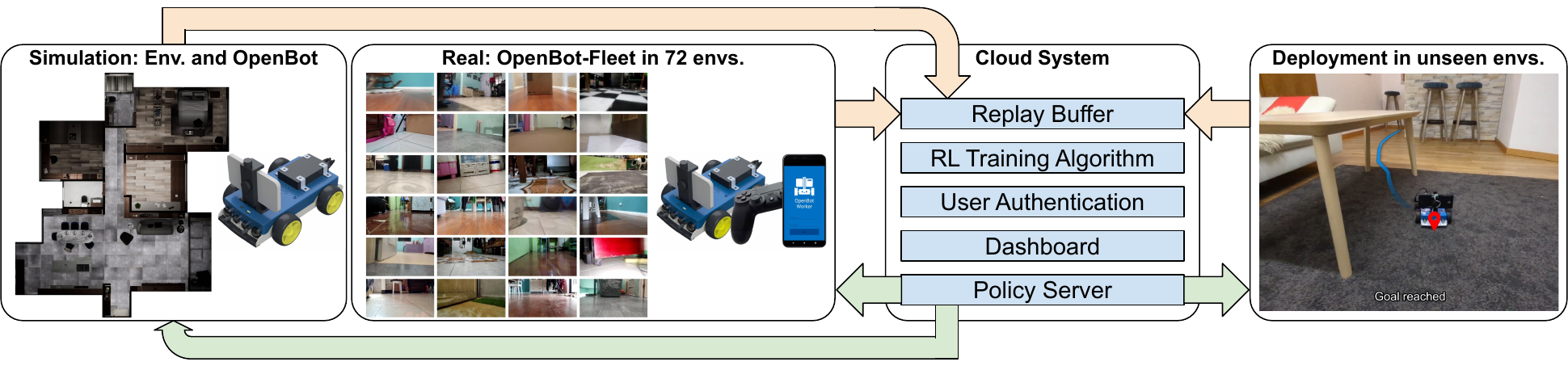}
 	\captionof{figure}{Overview of \systemName, a comprehensive cloud robotics system for navigation which supports both simulated agents and real robots. Our wheeled robots are robust, low-cost, and use a smartphone for sensing and compute. A smartphone app allows for teleoperation and control by a navigation policy, and secure communication with the cloud system to upload robot experience and receive policies. We train a self-supervised perception backbone using data from 72 robots and pretrain a controller in simulation. The full navigation policy is evaluated in unseen real-world environments.}\label{fig:intro}
\end{center}
}]

\begin{abstract}
We introduce~\systemName, a comprehensive open-source cloud robotics system for navigation.~\systemName uses smartphones for sensing, local compute and communication, Google Firebase for secure cloud storage and off-board compute, and a robust yet low-cost wheeled robot to act in real-world environments. The robots collect task data and upload it to the cloud where navigation policies can be learned either offline or online and can then be sent back to the robot fleet. In our experiments we distribute 72 robots to a crowd of workers who operate them in homes, and show that~\systemName can learn robust navigation policies that generalize to unseen homes with $>$80\% success rate.~\systemName represents a significant step forward in cloud robotics, making it possible to deploy large continually learning robot fleets in a cost-effective and scalable manner. All materials can be found at \url{https://www.openbot.org/}.
\end{abstract}

\section{Introduction} \label{sec:introduction}
Learning high-performance policies for robot fleets deployed in diverse and changing environments is often difficult because of the mismatch between robots and environments used for training vs. deployment. Cloud robotics systems~\cite{kehoe2015survey, wan2016cloud, saha2018comprehensive} can potentially address this issue by allowing deployed robots to collect and upload experience, improving the policy using this data, and sending the improved policy back to the robot fleet. However, while many open-source cloud robotics systems exist for different tasks~\cite{paull2017duckietown, pickem2017robotarium, pmlr-v176-bauer22a, dasari2019robonet}, almost none support the full policy improvement loop because of missing components like continuous policy deployment or experience data collection.

Offline learning (\eg, pretrained backbone networks for perception~\cite{deng2009imagenet, zeng2020transporter}) can reduce the amount of online robot experience needed to train high-performance policies \cite{herzog2023deep, shridhar2021cliport}. However, training robotics backbone networks is challenging because of the expenses associated with owning robots and operating them to act safely in real-world environments. Currently only large companies like Google and Tesla are able to operate large fleets of robots to acquire (private) datasets or implement the robot policy improvement loop at scale \cite{kalashnikov2018qt, grigorescu2020survey, brohan2022rt, wu2023autonomy}.

In this paper, we focus on point-goal navigation~\cite{anderson2018navigation} and present a comprehensive cloud robotics system
that enables the full policy improvement loop and tackles the challenges of cost and scalability (see \Cref{fig:intro}). To the best of our knowledge, ours is the first open-source distributed robot navigation system that can collect large-scale experience in the real world and continuously improve and deploy a policy across a fleet of robots.

We adapt OpenBot \cite{mueller2021openbot} to develop a wheeled robot designed for mass production and ease of use by crowd workers in home environments (see \Cref{fig:openbot_real}). Computation, sensing, and communication are done through a smartphone, leveraging the continuously improving capabilities of modern smartphones. We implement the full policy improvement software stack for both simulation (simulated environments, robot model, sensors) and the real world (cloud back end, smartphone app for robot control and sensor data logging) in a common learning framework. We demonstrate low-cost and scalable deployment by distributing our smartphone app and robot to a crowd of 72 workers, who provide the smartphone and home environment for learning point-goal navigation. The navigation policy consists of a perception module trained on data from this deployment and a control policy pretrained in simulation. This policy already achieves a success rate of 60\% in unseen test environments compared to 15\% by the best baseline. Online learning in these environments improves the success rate of our policy to 82.5\%. 

We make the following contributions:
(1) A robust, low-cost wheeled robot with a companion smartphone app that supports robot control through teleoperation or a policy, user authentication, sensor data logging, and secure communication with the cloud back end. %
(2) A cloud back end with experience replay buffer, policy learning framework, and a dashboard. %
(3) System validation through experiments where we deploy a large fleet of robots to a crowd of workers to learn point-goal navigation policies that generalize to unseen environments and significantly outperform baselines. %

\section{Related Work}
\label{sec:related_work}
\mypara{Robot Learning Systems} Recent advances in robot learning, particularly real-world deep reinforcement learning (RL), have significantly improved robot performance in locomotion~\cite{ibarz2021train, smith2022walk, Gandhi207IROS}, navigation~\cite{gattu2022autonomous, levine2023learning}, object manipulation~\cite{tian2019manipulation, kalashnikov2018qt, pmlr-v164-lu22a, herzog2023deep}, and grasping~\cite{pinto2016supersizing, Levine2018IJRR}. Despite the algorithmic advancements, the main challenges in robotics-oriented deep RL remain generalization and sample complexity~\cite{dasari2019robonet, ibarz2021train, pmlr-v164-lu22a}. To tackle these challenges, many researchers have focused on collecting large-scale datasets of multi-modal observations from multiple robots~\cite{pinto2016supersizing, Levine2018IJRR, Kalashnikov2018CoRL, kalashnikov2018qt, dasari2019robonet, brohan2022rt, reed2022Generalist}.
Additionally, the use of transformer models in combination with large-scale datasets has led to improved generalization performance~\cite{brohan2022rt, reed2022Generalist}. For example, RoboNet~\cite{dasari2019robonet} and Gupta \etal~\cite{gupta2018robot} focus on vision-based manipulation. The former demonstrates that offline training of a perception backbone network with such a dataset enables finetuning for the target robot that is more sample efficient than robot-specific policy learning. The latter shows that diversity of the data collection conditions (environment, robot, etc.) is important for generalization. However, these systems are expensive and not fully open-source and reproducible (see~\Cref{tab:system_comparison} for details). In contrast,~\systemName is affordable and completely open-source. In addition, while our experiments are designed for system validation, we observe similar effects with regard to sample efficiency and generalization.

\mypara{Affordable Systems and Large-Scale Datasets} Researchers have introduced several systems for data collection, policy training and benchmarking~\cite{ahn2019robel, yang2019replab, mueller2021openbot, foehn2022agilicious}. However, to the best of our knowledge,~\systemName is the only system that implements and open-sources the full policy improvement loop: experience data logging from a distributed fleet, secure cloud storage, learning framework, and policy deployment to the fleet.

\mypara{Point-goal Navigation} A prominent approach to solve this problem is SLAM~\cite{blochliger2018topomap} with optional visual landmark detectors and pose estimators~\cite{frintrop2008attentional, object2015occlusion}. Recent map-free algorithms try to address challenges arising from dynamic environments and textureless scenes by learning through self-collision with the environment~\cite{kahn2018self} or utilizing a transformer with the soft actor-critic algorithm~\cite{huang2023goal}. Note that the focus of this paper is a comprehensive cloud robotics system capable of continual policy improvement rather than the development of a novel point-goal navigation algorithm.

\section{System}
\label{sec:system}

We summarize our system below and provide a more detailed description of each component in the \supp.

\subsection{Robot Fleet}
\label{sec:fleet}

Each worker is given a wheeled robot, a remote control, and account credentials for logging into the smartphone app. 

\begin{figure}[!htb]
    \centering
    \includegraphics[width=\columnwidth]{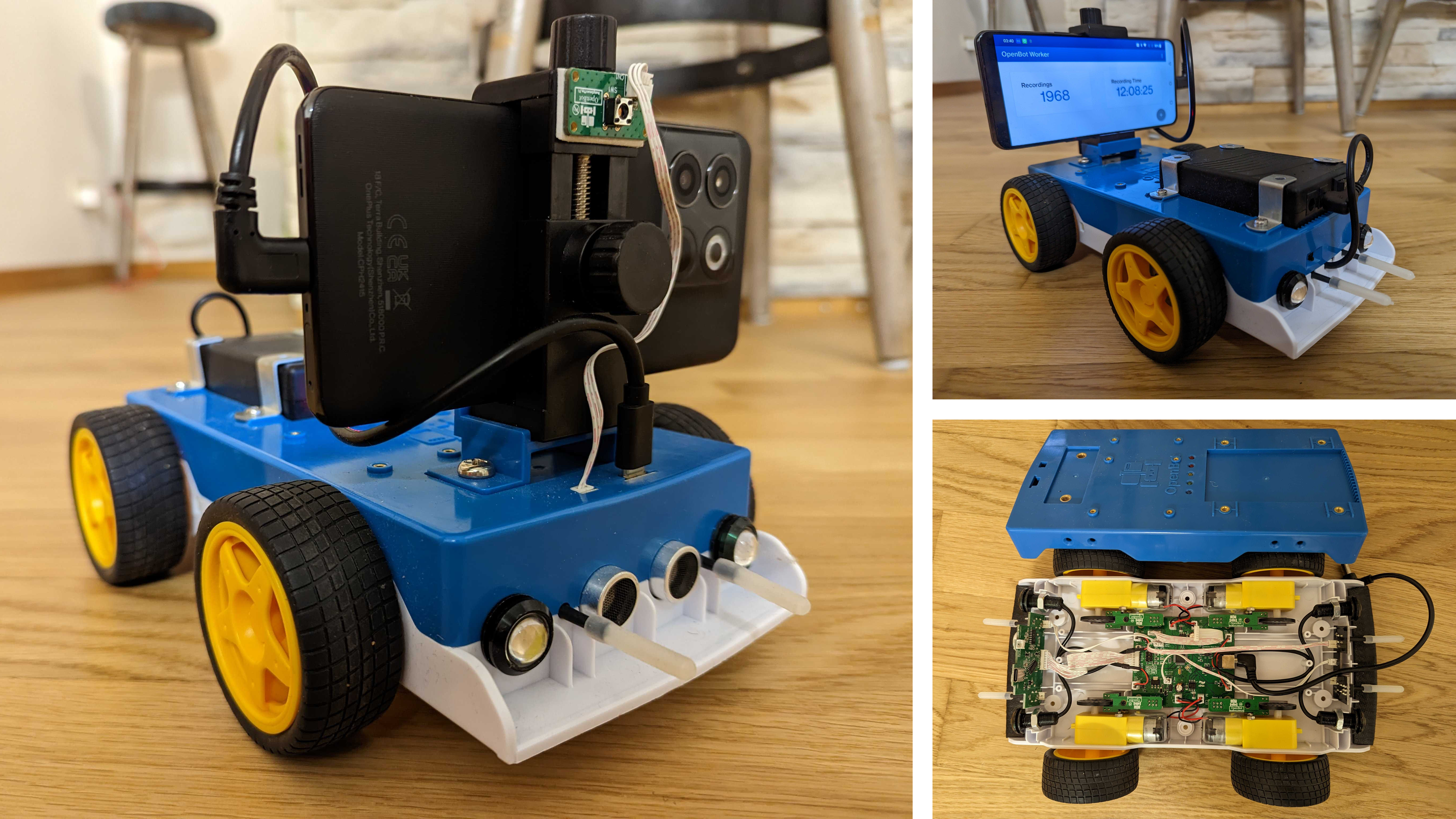}
    \caption{Our robot with a connected smartphone, adapted from OpenBot~\cite{mueller2021openbot} for shipping to and use by crowd workers. In contrast to the original DIY version of OpenBot, it has a robust injection-molded plastic shell, integrated electronics (visible in the bottom-right image), and improved cable routing.}
    \label{fig:openbot_real}
\end{figure}

\mypara{Robot}
The robot is based on the DIY OpenBot~\cite{mueller2021openbot} with several improvements vital for large-scale deployment while maintaining the low cost. The DIY OpenBot is 3D-printed and uses low-cost components like TT-motors with optical wheel encoders and an Arduino Nano. Building it is error-prone and time-consuming due to 3D-printing and manual tasks like wiring and soldering. To ensure durability and suitability for mass production, shipping, and repeated data collection with potential collisions, we have designed a robust injection mold plastic shell, fully integrated electronics with ATmega328 MCU, status LEDs and several sensors: wheel encoders, sonar, bump, and battery voltage (see~\Cref{fig:openbot_real}). A smartphone with our app for cloud robotics can easily be connected to the robot via a USB-C cable. Thus, our robot design addresses the challenges of cost and scalability encountered in cloud robotics.

\label{sec:robot}

\begin{figure}[!htb]
    \centering
    \includegraphics[width=\columnwidth]{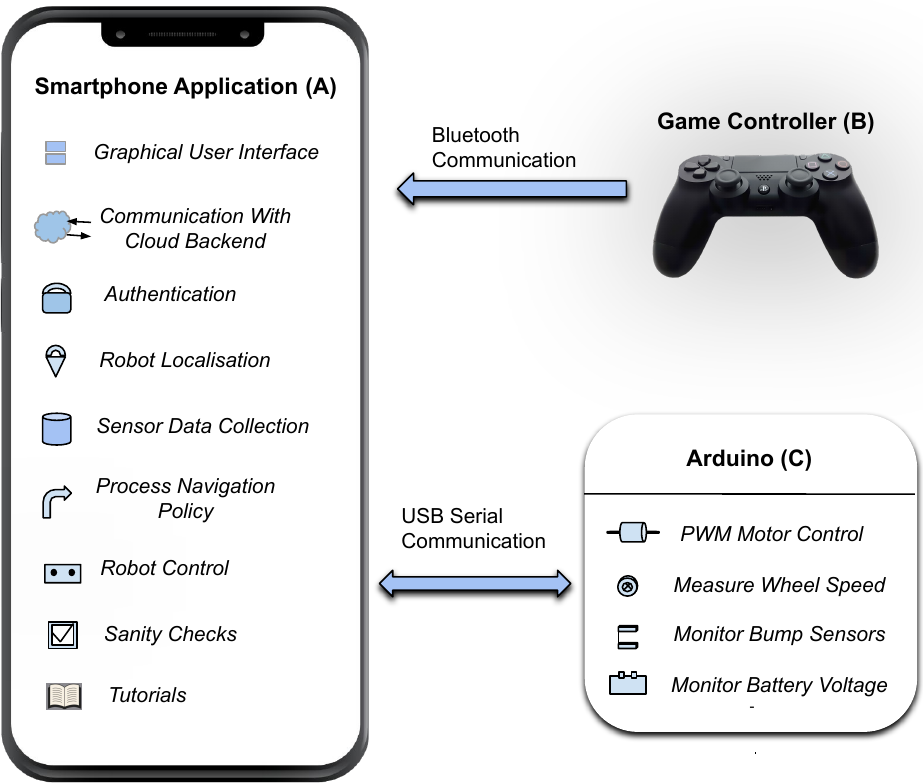}
    \caption{Features of our app which runs on the smartphone connected to each robot.
    It communicates with the robot via a two-way serial link to send controls to it (\eg, while executing a control policy) and to receive sensor data for logging. A Bluetooth game controller can be used for teleoperation.}
    \label{fig:fleet_overview}
\end{figure}

\mypara{Smartphone App}
\label{sec:app}
Our smartphone app allows robot control through game controller commands or navigation policies after user login. It also logs sensor data for each episode and securely communicates experience data and policy weights to and from the cloud back end.~\Cref{fig:fleet_overview} provides an overview of all app features. Detailed descriptions, an app-flow diagram, and screenshots can be found in the \supp.

\subsection{Policy Learning}
\label{sec:policy_learning}

Our system supports both offline learning, based on a dataset collected with the robot fleet, and online learning where a policy is continuously updated and redeployed based on continuously collected experience. In our method, we exploit both capabilities: (1)~We train a perception backbone with data collected by crowd workers using self-supervised learning. (2)~We update a control policy using reinforcement learning with real-world experience. 

\mypara{Offline Learning}
\label{sec:offline_learning}
Our framework supports training perception backbones on data collected with the fleet. We have implemented a lightweight but effective perception backbone based on U-Net \cite{ronneberger2015u}. Specifically, we train a floor segmentation network on the data collected by the robot fleet in a self-supervised fashion. We then sample points along the boundary as input representation for a control policy pretrained in simulation. This efficient and abstract representation allows for good sim-to-real transfer \cite{muller2018driving}. We have also implemented a pipeline for training end-to-end navigation policies using imitation learning. Please refer to \Cref{sec:baselines} (Baselines) for more details.

\mypara{Online Learning}
\label{sec:online_learning}
Our system also supports online learning with a fleet of robots deployed in the real world. This requires a cloud system for collecting experience from the robot fleet, updating the policy, and pushing it back to the fleet. To this end, we use Google Firebase~\cite{google_firebase} secure cloud storage.

Our learning framework is based on TensorFlow Agents~\cite{TFAgents} and the architecture is shown in Fig. 1-C of the \supp. The \emph{driver} uses the current \emph{collect policy} to execute an episode with the agent in the environment, and feeds the output \emph{trajectory} to the \emph{replay buffer}. Each \emph{trajectory} contains a tuple of consecutive observations, action, reward, and additional meta-data, \eg, if a step caused the end of an episode. Note that the data collection and learning pipeline are asynchronous, so multiple agents can collect experience in different environments at the same time. 

We implement a simulation driver (see~\Cref{sec:simulation}) and a real-world driver, both of which can interact with the common replay buffer and policy update framework. The real-world driver leverages TensorFlow Lite~\cite{tf_lite} to run our TensorFlow policies efficiently on smartphones. While executing an episode, it records a log file containing the observations, policy action, and additional information like the robot pose relative to the target, for each step. Once the target is set by the worker as described in~\Cref{sec:data_collection}, we use ARCore~\cite{du2020depthlab, ar_core} to track the robot pose. After the episode is over, this log is uploaded to the cloud storage. The learning framework automatically retrieves it, converts it into the \emph{trajectory} format, and adds it to the replay buffer. This design decision of not computing the reward directly on the smartphone allows greater flexibility in experimenting with the reward function later.

\subsection{Simulation Platforms}
\label{sec:simulation}

We have integrated two simulators into our system: ProceduralSim, a fast procedural simulator, and SpearSim \cite{spear}, a simulator based on UnrealEngine with photo-realistic indoor environments. We use ProceduralSim for pretraining a control policy using reinforcement learning. This learned controller can then be deployed on a real robot in conjunction with a perception backbone (see Section IV-A of the \supp) and can be further finetuned with our online learning framework. We use SpearSim mainly for evaluating our complete navigation stack before real-world deployment. Please refer to the \supp for more details.

\subsection{Comparison to Other Systems} \label{sec:system_comparison}
Table~\ref{tab:system_comparison} presents a comparison of our proposed distributed learning system to other works. While the largest deployed fleets of robots in industry, \ie, autonomous cars and robots for warehouse logistics and last mile delivery, require good navigation policies, most existing research systems for large-scale learning focus on manipulation. In contrast, our system focuses on navigation. The small cost compared to other systems also enables much larger deployments. Our system is designed to be deployed in the wild and the user interface through a smartphone app does not require robotics experts for operation. Other systems are usually deployed in controlled lab settings and require expert supervision. To reproduce a system, code and instructions for re-creating the system (including the robots) need to be available. Similarly, data, models and code need to be released for model reproducibility. While other works only release one or the other, we open-source and release all of it.

\begin{table}[!htb]
\setlength{\tabcolsep}{2pt}
\centering
\begin{tabular}{c|ccc|cc|cc|cc}
\toprule
\multirow{2}{*}{System} & \multicolumn{3}{c|}{Robots} & \multicolumn{2}{c|}{Envs} & \multicolumn{2}{c|}{Tasks} & \multicolumn{2}{c}{Reproducible} \\
 & Type & Cost [\$]& \# & Type & \# & Man. & Nav. & ~Sys. & Mod. \\
\midrule
Gupta \etal~\cite{gupta2018robot} & MM & 3,000  & 5  & W & 6       & \YesV  & \NoX & \YesV & \NoX  \\
RoboNet~\cite{dasari2019robonet}  & SM & 26,000 & 7  & L & 4       & \YesV  & \NoX  & \NoX  & \YesV \\
REPLAB~\cite{yang2019replab}      & SM & 3,000  & 2  & L & 3       & \YesV  & \NoX & \YesV & \NoX  \\
QT-Opt~\cite{kalashnikov2018qt}   & SM & 25,000 & 7  & L & 7       & \YesV  & \NoX & \NoX  & \NoX  \\
RT-1~\cite{brohan2022rt}          & MM & N/A*   & 13 & L & $\ge$8 & \YesV & \YesV & \NoX  & \YesV \\
\midrule
Ours                              & GV & 100     & 72 & W & 72      & \NoX & \YesV  & \YesV & \YesV \\
\bottomrule

\end{tabular}
\caption{Comparison of our proposed multi-robot distributed learning system to other similar works. Robot types are denoted by MM (Mobile Manipulator), SM (Static Manipulator), and GV (Ground Vehicle). Environment types are denoted by W (in-the-wild deployment) and L (controlled lab setting). Manipulation, Navigation, System, and Model are abbreviated with the first three letters. *Robots not available for purchase.}
\label{tab:system_comparison}
\end{table}

\section{Methodology}
\label{sec:methods}

We implement a robotic navigation system that can drive a robot from A to B while avoiding obstacles. %

\subsection{Large-Scale Data Collection}
\label{sec:data_collection}
We collect large-scale driving data with the help of crowd workers. The task is to drive from A to B, taking the shortest path while avoiding obstacles. The data collected consists of image-control pairs and target locations. We use this data for (1) learning a conditional imitation learning~\cite{codevilla2018end} baseline which regresses controls end-to-end from images and relative target poses, and (2) training our self-supervised perception backbone from images.

The data collection process involved 72 crowd workers using Bluetooth controllers and smartphones attached to the robot bodies to record data in diverse natural indoor home environments such as living rooms, bedrooms, and kitchens. The data contains images collected by the smartphone camera, synchronized with controls supplied by the human operators and state estimation computed with ARCore. The complete dataset consists of 15,642 recordings and over 1M frames with a variable frame rate of up 10 frames per second depending on the smartphone capability. 

Participants were given specific instructions to ensure the quality of the data collected. They were instructed to place the robot on the ground before commencing the recording, to drive smoothly from the starting point to the goal without stopping, using the shortest path while maneuvering around obstacles such as furniture, walls, and naturally placed household items. Also, to ensure that the trajectory length is at least 2 meters and to prevent the robot from being picked up during the recording, an automatic check was put in place. To ensure a diverse set of data, participants were instructed to change the start and goal locations at least after every 5 to 10 episodes. They were also instructed to avoid certain behaviors during the episodes such as driving in circles or zig-zagging, and stopping the robot during the task unless they had reached the end point or the robot had collided. To ensure privacy, participants were also instructed to only record in their own home environment, avoid people's faces and actively review each recording before submitting it. Despite best efforts, the data quality varies based on worker experience and motivation, robot, and environment condition.

\subsection{Learning Perception}
Our collected dataset is large, diverse, and suitable for training and evaluating various perception modules for the robot. We train two modules. (1) A end-to-end baseline (see~\Cref{sec:baselines}) that regresses the action to be applied for the next step from a given RGB image supervised by the synchronized teleoperation command. (2) For our system, a perception backbone that predicts the floor boundary from a given RGB image.

The perception backbone for the floor boundary prediction is trained in a scalable self-supervised manner which does not require human labelling. We generate an initial floor mask by applying a gradient threshold on the depth image generated by MiDaS~\cite{ranftl2020towards} from the RGB image and further refine the mask using GrabCut~\cite{rother2004grabcut}. For more details, see Section IV-A in the \supp. We use the more than 1M images collected by the crowd workers and an additional 12k images from SpearSim (to reduce the sim-to-real gap between evaluation in simulation and the real world). The resulting dataset contains 1,068,577 images in total and is split randomly into 90\% for training and 10\% for validation. The network is based on UNet~\cite{ronneberger2015u} and trained to segment the floor. The boundary is the uppermost pixel location of this mask for each image column, and is used as an observation representation for our navigation policy.  

\subsection{Learning Control}
We train a control policy using reinforcement learning. The policy is a residual network with respect to a unicycle model. We first pretrain in simulation using ground truth observations to collect a large amount of experience quickly. We then finetune the policy using the real robot fleet. 

\mypara{Unicycle Model}
By default, the robot moves towards the goal using a unicycle model:
\begin{equation}
    v_l = \frac{\cos{\alpha} - \sin{\alpha}}{\sqrt{2}} 
    \quad\text{and}\quad
    v_r = \frac{\cos{\alpha} + \sin{\alpha}}{\sqrt{2}} \,,
    \label{eq:unicycle}
\end{equation}
where $v_l$, $v_r$ denote the left and right motor commands, and $\alpha$ is the target heading angle w.r.t. the robot front. The learnt policy predicts residual actions $\Delta v_l$, $\Delta v_r$ which are added to $v_l$, $v_r$ to get the final motor commands.

\mypara{Learning Algorithm}
We use the popular off-policy RL algorithm Soft Actor-Critic (SAC)~\cite{Haarnoja2018}. Thanks to the use of a large replay buffer where past experiences are stored and re-used to train the policy, SAC is usually more sample efficient than on-policy RL algorithms like Proximal Policy Optimization (PPO) ~\cite{schulman2017proximal}. Additionally, SAC has an entropy bonus that encourages exploration. 

\mypara{Training Process}
Once a training session is started, the perception backbone, control policy and training meta-data are uploaded to our cloud back end. The worker's app gets notified and the worker can collect an episode recording and upload it. The uploaded recording is stored in the cloud and immediately passed on to the training session, where we add it to the replay buffer and update the RL policy. The updated policy is published to the workers and the cycle repeats.

\subsection{Real-world deployment and evaluation}
\begin{figure}[!htb]
\centering
\includegraphics[width=\columnwidth]{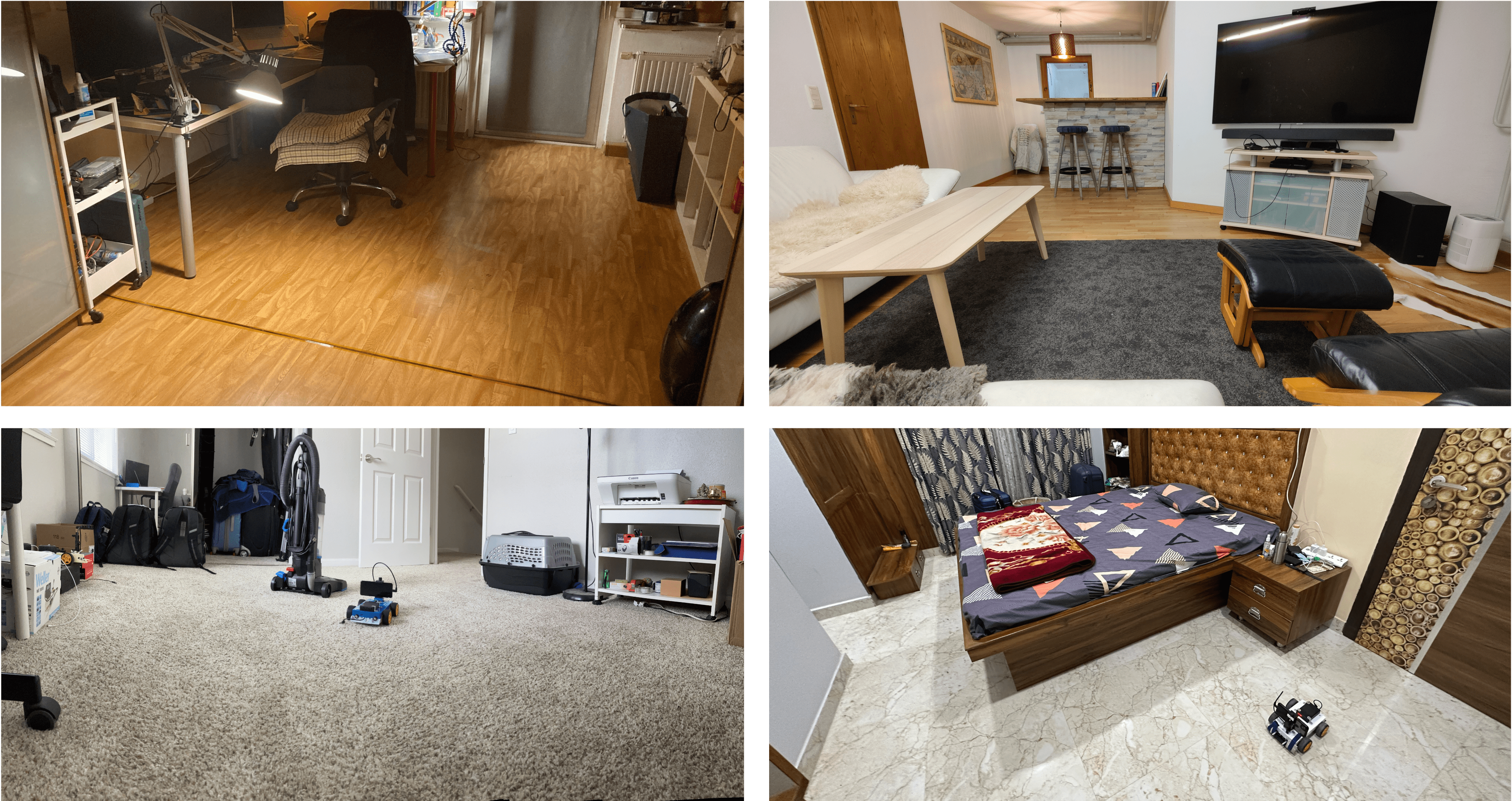}
    \caption{Unseen real-world environments used for evaluation.}
    \label{fig:real_envs_eval}
\end{figure}

We deploy our policy to real robots in different unseen environments (see \Cref{fig:real_envs_eval}) to evaluate both the zero-shot and few-shot performance. We compare our approach to several baselines.

\mypara{Task Definition}
\label{sec:task}
We evaluate our system on the standard and yet challenging task of point-goal navigation~\cite{anderson2018navigation}. Here, the goal is to learn a navigation policy that automatically drives the robot (without collisions) to a user-defined target in an indoor environment. We have chosen point-goal navigation as the optimal task for evaluating our system due to several reasons. Firstly, most smartphones support ARCore which can be used for state estimation, defining the goal position, and computing the reward for the reinforcement learning algorithm. Secondly, the point-goal navigation task is simple to define, yet a challenging problem to solve. This task is also useful for practical applications in real-world environments, \eg, autonomous vacuum cleaners or home assistants. Finally, this task is well-suited for evaluating the effectiveness of our proposed approach in solving navigation problems at scale.

\mypara{Evaluation Protocol and Metrics}
We regard a policy rollout as successful if the robot reaches the target in a specified time without collision. A collision is either automatically detected by the bump sensors or identified by user intervention.
We apply three metrics to evaluate a policy: success rate, distance to the goal at the final time step, and collision rate. An episode ends when either the goal is reached, the robot collides, or the maximum number of steps is reached (timeout). On the real robot, episodes can also end due to ARCore failures.

\mypara{Baselines}
\label{sec:baselines}
We implement a goal-conditioned imitation learning network~\cite{codevilla2018end} with PilotNet backbone \cite{bojarski2016end} that regresses controls end-to-end based on ARCore goal and images (CIL \cite{codevilla2018end} - PilotNet \cite{bojarski2016end}). For this baseline, the data collected by the crowd workers is first split into a training set and a validation set, with 65 workers contributing 14,141 recordings and 957,434 frames to the training set, and 7 workers contributing 1,501 recordings and 98,198 frames to the validation set. The data is then filtered to remove bad recordings and frames where no control signal was supplied (\eg, robot is stationary) resulting in 590,465 image-control pairs in the training set and 60,291 image-control pairs in the validation set. This baseline provides a comparison for the performance of our proposed method to behaviour cloning. 

We also implement two other baselines, representing different observation representations: (1) \textit{MobileNetV2}: image features extracted with MobileNetV2 \cite{howard2017mobilenets} trained on ImageNet \cite{deng2009imagenet}, and (2) \textit{DepthRays}: the depth values of a pencil of 11 rays parallel to the ground and originating at the camera center, normalized in the $[0, 1]$ range. Both representations are used to train a navigation policy with reinforcement learning in simulation in the same fashion as for our approach. The simulation training phase uses the ground truth RGB rendering and depth image respectively, similar to the simulation training phase of our policy, which uses the ground truth floor boundary. The real world deployment and evaluation phase uses the real RGB image and the the real depth image generated by ARCore \cite{ar_core, du2020depthlab}.

\begin{table*}[!htb]
\setlength{\tabcolsep}{5pt}
\centering
\begin{tabular}{l|ccc|ccc|ccc|ccc|ccc}
\toprule
      & \multicolumn{3}{c|}{Environment 1} & \multicolumn{3}{c|}{Environment 2} & \multicolumn{3}{c|}{Environment 3} & \multicolumn{3}{c|}{Environment 4} & \multicolumn{3}{c}{Average} \\
      & SR\,$\uparrow$& GD\,$\downarrow$ & CR\,$\downarrow$ & SR\,$\uparrow$ & GD\,$\downarrow$ & CR\,$\downarrow$ & SR\,$\uparrow$ & GD\,$\downarrow$ & CR\,$\downarrow$ & SR\,$\uparrow$ & GD\,$\downarrow$ & CR\,$\downarrow$ & SR\,$\uparrow$ & GD\,$\downarrow$ & CR\,$\downarrow$\\
\midrule
CIL \cite{codevilla2018end} - PilotNet \cite{bojarski2016end} & 0 & 1.07 & 100 & 10 & 0.49 & 80 & 0 & 1.86 & 100 & 0 & 1.68 & 100 & 2.5 & 1.28 & 95.0  \\
\midrule
MobileNetV2 \cite{howard2017mobilenets} (zero-shot) & 0 & 0.27 & 70 & 10 & 1.14 & 40 & 0 & 2.73 & 70 & 10 & 2.07 & 90 & 5.0 & 1.55 & 67.5  \\
DepthRays \cite{du2020depthlab} (zero-shot) & 30 & 0.80 & 70 & 30 & 0.81 & 60 & 0 & 1.87 & 100 & 0 & 2.19 & 100 & 15.0 & 1.42 & 82.5  \\
Ours (zero-shot) & \underline{50} & \underline{0.31} & \underline{40} & \underline{60} & \underline{0.51} & \underline{30} & \underline{50} & \underline{0.79} & \underline{50} & \underline{80} & \underline{0.09} & \underline{20} & \underline{60.0} & \underline{0.43} & \underline{35.0} \\
\midrule
Ours ($\leq$\,200 episodes) & \textbf{80} & \textbf{0.20} & \textbf{20} & \textbf{80} & \textbf{0.02} & \textbf{10} & \textbf{80} & \textbf{0.27} & \textbf{20} & \textbf{90} & \textbf{0.02} & \textbf{10} & \textbf{82.5} & \textbf{0.13} & \textbf{15.0} \\
\bottomrule
\end{tabular}
\caption{Point-goal navigation results with real robot in unseen environments. \emph{SR}: success rate [\%], \emph{GD}: goal distance at the end of the episode [m], \emph{CR}: collision rate [\%]. \textbf{Bold} for best, \underline{underlined} for second-best. Models are pretrained for 20k episodes in ProceduralSim and use a history length of 5. All models are evaluated for 10 trials and results are averaged. We finetune models until they achieve a success rate of 100\% in 10 consecutive training episodes or reach 200 episodes.} \label{tab:exps_real}
\end{table*}

\section{Results}

\subsection{Zero-shot generalization}
Our evaluation on real robots in four unseen environments is shown in \Cref{tab:exps_real}. Our method achieves superior zero-shot performance with a success rate (SR) of at least 50\% and up to 80\%. The next best method is DepthRays, achieving 30\% in two of the environments. However, it completely fails in the other two (0\% SR). This is probably due to the noisy and unreliable depth compared to simulation. MobileNetV2 performs even worse with 0\% SR in all environments but one. We conjecture that this due to the distribution mismatch -- the camera mounted on the robot is very low to the ground and images contain mostly floor and furniture whereas most images in ImageNet are captured by humans and focus on diverse objects in various settings. The behaviour cloning baseline, CIL-PilotNet, also performs poorly despite the additional supervision from the control labels by the crowd workers. Behaviour cloning is sensitive to bad samples \cite{sasaki2021behavioral}, and not all of the workers can be considered experts. In addition, they may have different driving styles and optimize for task completion rather than trajectory quality. 

\subsection{Few-shot adaptation}
We also investigate the potential of our method to improve performance with online adaption in the real world. To this end, we deploy our policy in different unseen environments and finetune the policy on up to 200 episodes using our online learning system. Our method achieves at least 80\% and up to 90\% success rate across all environments with less than 200 episodes of online learning. Note that state estimation by ARCore is noisy which can lead to unreachable goals making a success rate of 100\% challenging to impossible. 

We also experimentally confirm that the finetuned policies do not overfit to the environments. For example, the policy finetuned in environment 1, still achieves 70\% success rate in environment 4. We also compare the adaption performance of our method to the baselines with up 4,000 episodes of finetuning per policy using the photo-realistic simulator SpearSim. In summary, only our method is able to quickly adapt and achieve high performance. %

\begin{table}[!htb]
    \centering
    \setlength{\tabcolsep}{5pt}
    \begin{tabular}{c|cc|ccc}
        \toprule
        \# of envs. & mIOU\,$\uparrow$ & MAE\,$\downarrow$ & SR\,$\uparrow$ & GD\,$\downarrow$ & CR\,$\downarrow$ \\
        \midrule
        1 & 90.95 & 4.58 & 50 & 0.69 & 50 \\
        6 & 91.94 & 4.01 & 70 & 0.44 & 30 \\
        72 & 93.49 & 2.95 & 85 & 0.095 & 15 \\
        \bottomrule
    \end{tabular}
    \caption{Ablation study of the perception backbone. We train with data from 1, 6, and 72 environments and show that diversity and scale improve both the accuracy of the perception backbone and the performance of the navigation policy. \emph{mIOU}: mean intersection over union for floor segmentation [\%], \emph{MAE}: mean average error for floor boundary, \emph{SR}: success rate [\%], \emph{GD}: goal distance at the end of the episode [m], \emph{CR}: collision rate [\%].}
    \label{tab:results}
\end{table}

\subsection{Ablation studies}
To study the impact of diversity and scale, we train the perception backbone on data from 1, 6, and 72 workers. In the case of a single environment, we pick one that is relatively similar to the test environment used for backbone evaluation and sample 217 recordings (the average number of recordings per environment) for training a perception backbone. Similarly, for the case of 6 environments we train a model on a total of 1302 recordings  (6 $\times$ 217) across 6 environments. Finally, we also train a model on all data from all 72 environments. We then evaluate the performance of the perception backbone on our test set and also evaluate the zero-shot task-performance in different novel environments. 

The results presented in Table \ref{tab:results} show the correlation between perception backbone metrics and zero-shot task performance. We observe a clear trend: as the diversity and quantity of data increases, the performance metrics improve. Specifically, the mIOU increased from 90.95\% for 1 environment to 93.49\% for 72 environments, indicating the effectiveness of learning at scale and in diverse environments. 

Furthermore, we evaluate the perception backbone in conjunction with our controller trained in simulation on two novel unseen real-world environments. The success rate (SR), collision rate (CR), and average remaining distance to the goal (GD) were measured across 20 trials. The results show that increasing the number of training environments leads to improved task performance. For instance, when trained with data from 72 environments, the system achieves a SR of 85\%, a CR of 15\%, and an average remaining distance to the goal of 0.095m.

These findings highlight the importance of both diversity and scale in training good navigation polices and demonstrates the usefulness of our cloud robotics system. The correlation between the perception backbone metrics and task performance confirms the effectiveness of our crowd-sourced data collection approach. Collecting data from a larger number of diverse environments not only improves the perception backbone performance but also enhances the system's overall performance in real-world scenarios. The presented results validate the effectiveness and impact of our system in achieving robust and reliable performance in cloud robotics applications. We provide further ablation studies for selecting the parameters of our approach in the \supp.

\section{CONCLUSION} 
\label{sec:conclusion}
We presented a comprehensive cloud robotics system that addresses cost and scalability challenges in learning navigation policies with robot fleets in real-world environments. It consists of a low-cost, plug-and-play robot equipped with a robust injection-molded plastic shell, integrated electronics, and a software stack that enables data collection and online learning for large fleets of robots. We demonstrated the effectiveness of our system through experiments with a fleet of 72 robots distributed to crowd workers. The resulting navigation policies exhibit good performance in unseen environments with less than 200 episodes of online learning. %

While the results are promising, there are some limitations. The robot's actuation is noisy, limiting zero-shot performance in new environments. ARCore, used for state estimation, exhibits drift and frequently loses tracking since it was not designed with robot navigation in mind. Beyond modeling the noisy robot actuation and improving state estimation, it would also be interesting to scale up to fleets beyond 1,000 and to continuously update both the perception backbone and the control policy.

A video with a short summary of this work can be found at \url{https://youtu.be/3tuKzbtdFcc}. The supplementary material which contains further implementation details is available at \url{https://t.ly/v9be9}.

\clearpage
\bibliographystyle{IEEEtran}
\bibliography{references}
\begin{appendices}
\section*{Appendices}

\begin{figure*}[!htb]
    \centering
    \includegraphics[width=\textwidth]{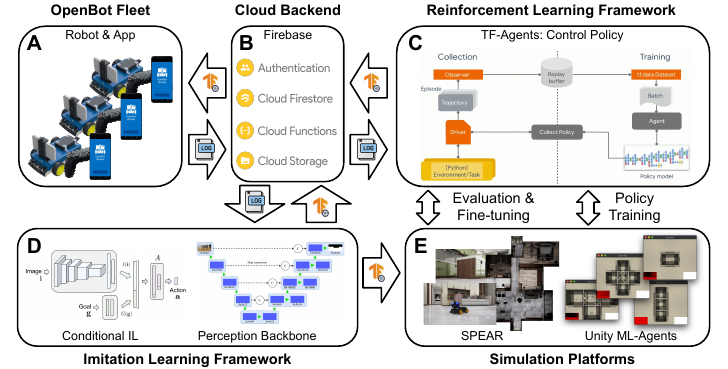}
    \caption{Overview of our system for collective learning with a fleet of mobile robots. The cloud backend (B) publishes recording requests that are observed by the robot fleet (A) via real-time listeners. After the successful upload of a recording, the backend either stores the data in a database for offline learning or notifies an online learning process to consume the data directly. When a new policy is available, it is automatically redeployed to the robot fleet. To validate our system, we use a large fleet of mobile robots to train a perception backbone using self-supervised learning and also implement goal-conditioned imitation learning \cite{codevilla2018end} based on PilotNet \cite{bojarski2016end} as a baseline (D). We have also built a fast procedural simulator (ProceduralSim) based on Unity ML Agents \cite{juliani2018unity} for pre-training navigation policies and integrated the photorealistic simulator SpearSim \cite{spear} to perform extensive evaluations before real-world deployment~(E). We pre-train a control policy in Unity based on privileged perception information using reinforcement learning. Both the perception backbone and pre-trained control policy are packaged into TensorFlow-lite models for deployment to the smartphones, which execute them to control the robots (A). Once deployed, the system can be further finetuned via online learning. Details are given in Section III of the main paper.}
    \label{fig:system_overview}
\end{figure*}

We depict an overview of our system in \Cref{fig:system_overview}. It consists of four main building blocks that we discuss in the following.

\section{Robot Fleet}
\label{sec:fleet}

Each worker is equipped with a robot body, a smartphone with our app installed, and a remote control; see \Cref{fig:system_overview}-A.

\subsection{Robot}
Our low-cost robot addresses the challenges of cost and scalability in cloud robotics. The robot is designed to be mass-produced and to be plug-and-play. It is built with low-cost components such as the TT-motors with optical wheel encoders and the ATmega328 MCU. The robot also includes sonar, bump, and battery voltage sensors for collision detection and power monitoring. To ensure its durability and suitability for mass production, the robot is equipped with a robust injection mold plastic shell and fully integrated electronics. A smartphone with our app for cloud robotics (refer to Section III-A of the main paper) can easily be connected to the robot via a USB-C cable. The images in Figure 3 of the main paper show the robot with a connected smartphone and the inside of the robot showcasing the integrated electronics.

\begin{figure*}[!htb]
\centering
\includegraphics[width=\textwidth]{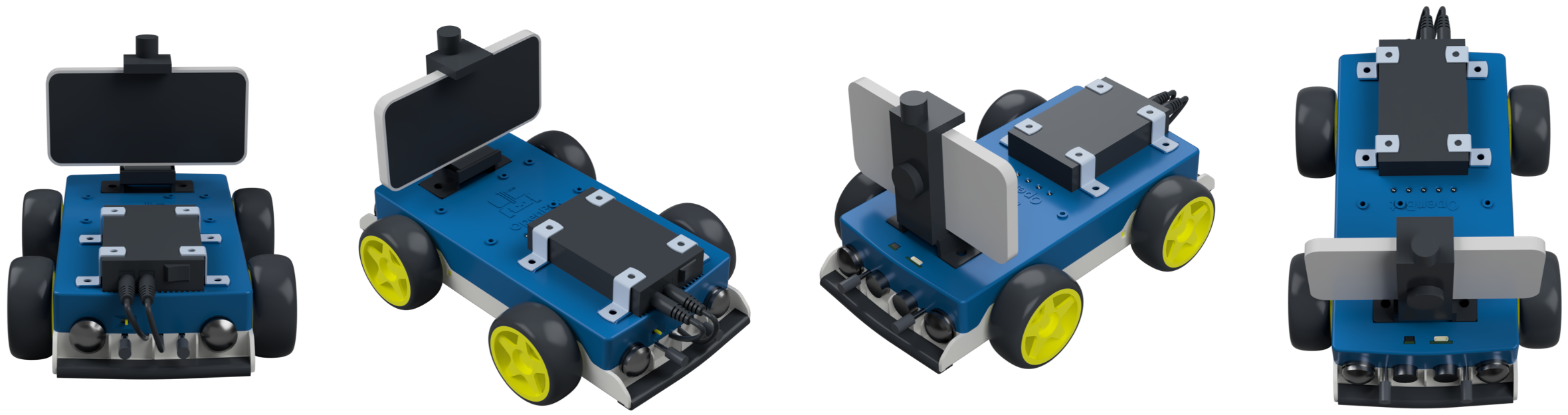}
    \caption{Renderings of the robot.}
    \label{fig:openbot_model}
\end{figure*}

\Cref{fig:openbot_model} shows renderings of the robot from various angles. The different views illustrate the robot's physical appearance, design, and functionalities.

\subsection{Smartphone App}
\label{sec:app_detail}
The smartphone app is responsible for communicating with the cloud backend, executing navigation policies, and controlling the robot. The flow of the app is shown in \Cref{fig:app_flow}. Upon opening the app, the user must enter their login credentials for verification by the cloud backend before getting to the home screen. The home screen displays information on the total number of executed recordings and the corresponding recording time (see top-left of \Cref{fig:app_screens_recording}).  

\subsubsection{Authentication}
When the app is first opened the user needs to enter their login credentials at the login screen. The credentials are then sent to the cloud backend for verification. This adds a layer of security and also helps with data access management.

\subsubsection{Sanity Checks}
\Cref{fig:app_screens_checks} shows a set of dedicated fragments that allow the user to perform some checks to make sure the robot works correctly and to ensure that the Bluetooth remote control is properly paired to their phone and working as expected. For the robot, the user can perform simple tests like checking the wheel motion and speed. The game controller checks allow the user to test the button mapping. 

\subsubsection{Tutorials}
We provide tutorials in the app that allow the users to learn about the recording tasks. The tutorial helps in correctly setting up the environment and describes the steps needed to perform a particular task. Data collection requires the completion of a tutorial  (refer to \Cref{fig:app_screens_instructions}) containing all the required instructions to properly operate the robot and a dummy recording (see \Cref{fig:app_screens_recording}) to become familiar with the process.  

\subsubsection{Recording Requests}
\begin{figure}[ht]
    \centering
    \includegraphics[width=\columnwidth]{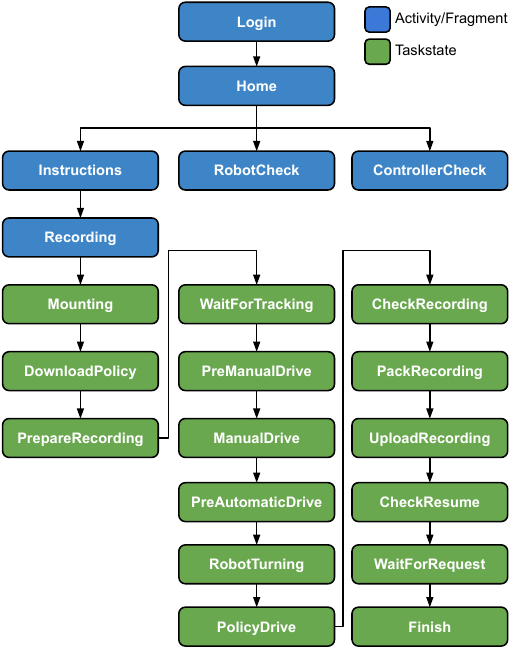}
    \caption{Flow of our app, showing the different fragments and all the steps to complete a task.}
    \label{fig:app_flow}
\end{figure}

\begin{figure*}[!htb]
\centering
\includegraphics[width=\textwidth]{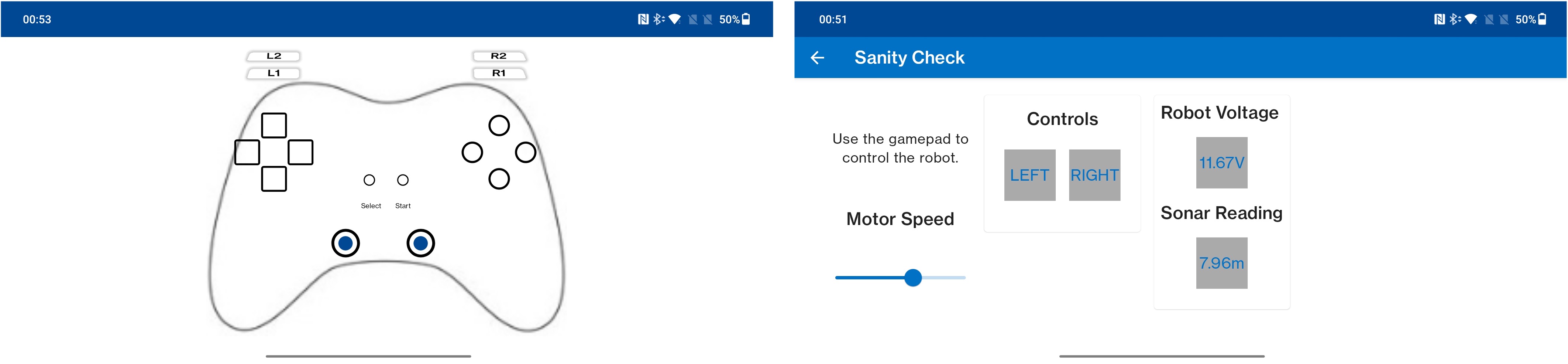}
    \caption{Screenshots of the sanity checks screens in the smartphone app.}
    \label{fig:app_screens_checks}
\end{figure*}

\begin{figure*}[!htb]
\centering
\includegraphics[width=\textwidth]{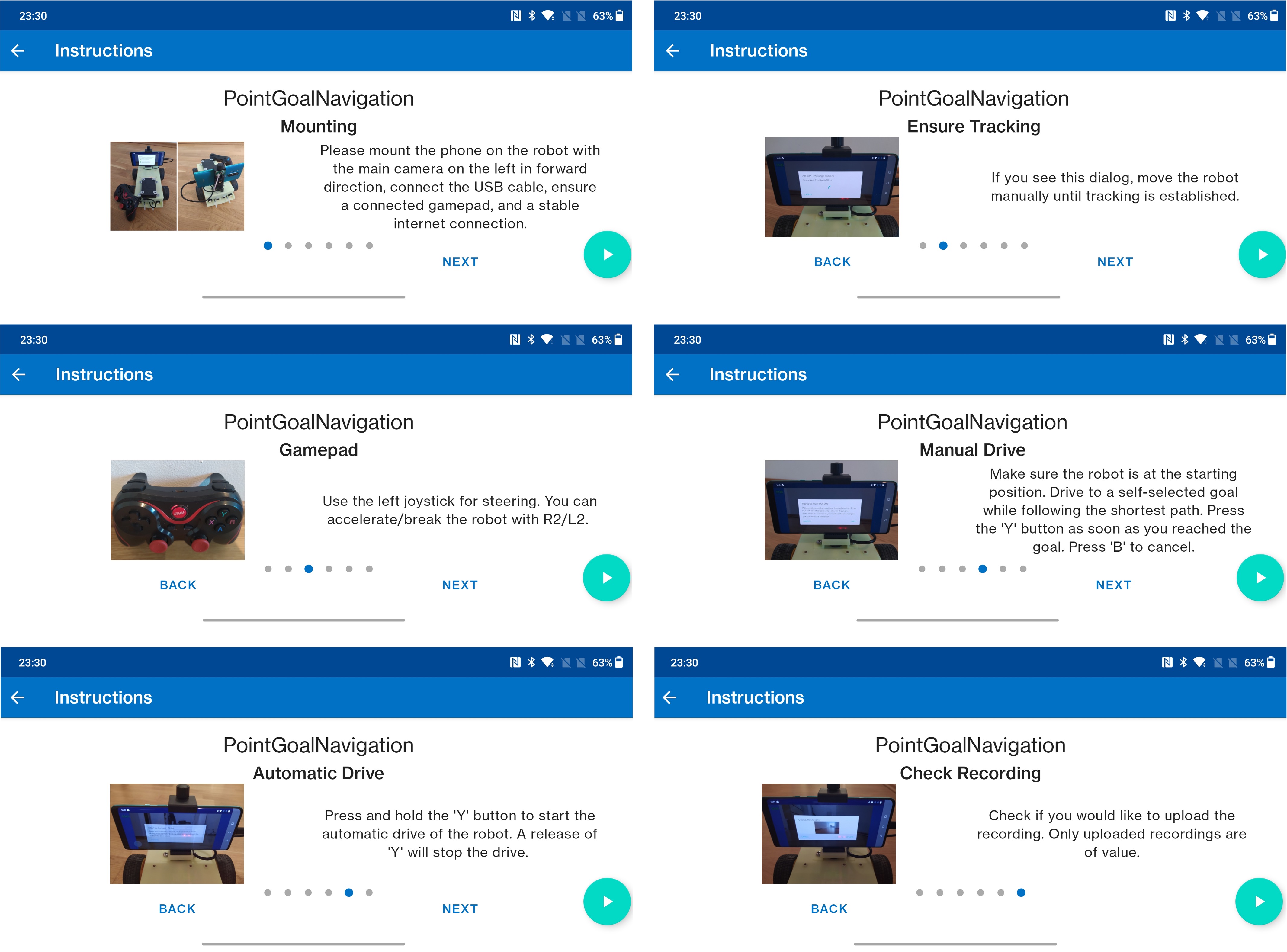}
    \caption{Screenshots of the instruction screens in the smartphone app.}
    \label{fig:app_screens_instructions}
\end{figure*}

\begin{figure*}[!htb]
\centering
\includegraphics[width=\textwidth]{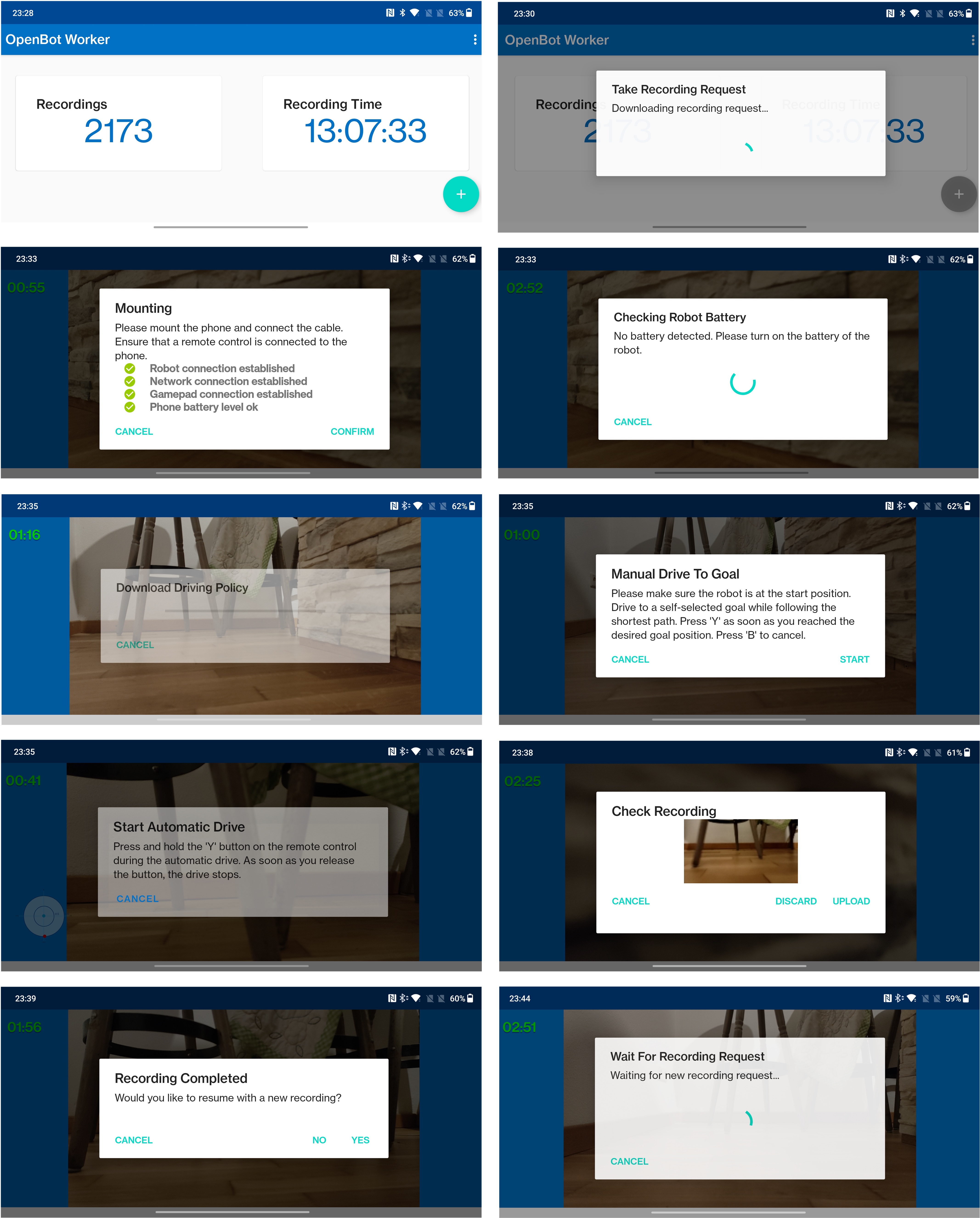}
    \caption{Screenshots of the recording screens in the smartphone app.}
    \label{fig:app_screens_recording}
\end{figure*}

The `+' button on the lower-right corner of the screen (see top-left of \Cref{fig:app_screens_recording}) informs the user about the availability of recording requests from the cloud backend. This button turns green when a recording request is available for the user. The user can accept the recording request and then needs to go through some checks to ensure that robot and phone are connected, a network connection exists, the remote control is properly connected and that the battery level of the phone and robot are sufficient. After these checks, the user can go through all of the different task states as indicated by the green check boxes in \Cref{fig:app_flow}. The corresponding app screens are shown in \Cref{fig:app_screens_recording}. The app utilizes the ARCore API for determining the relative position and orientation of the robot during Automatic Drive.

\subsubsection{Stop Reasons}
\begin{figure*}[!htb]
\centering
\includegraphics[width=\textwidth]{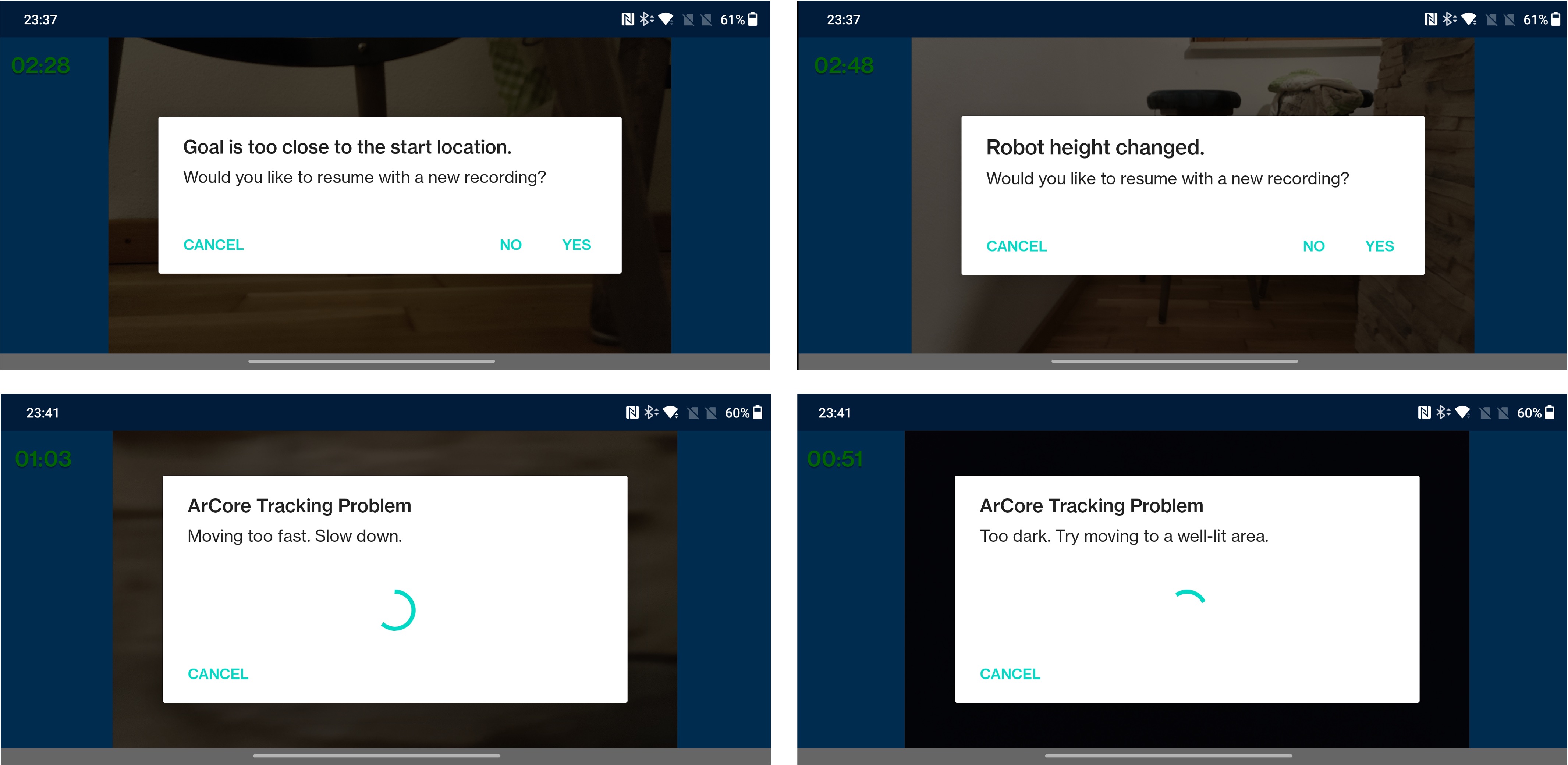}
    \caption{Screenshots of the failure mode screens in the smartphone app.}
    \label{fig:app_screens_failure}
\end{figure*}

\begin{figure*}[!htb]
\centering
\includegraphics[width=\textwidth]{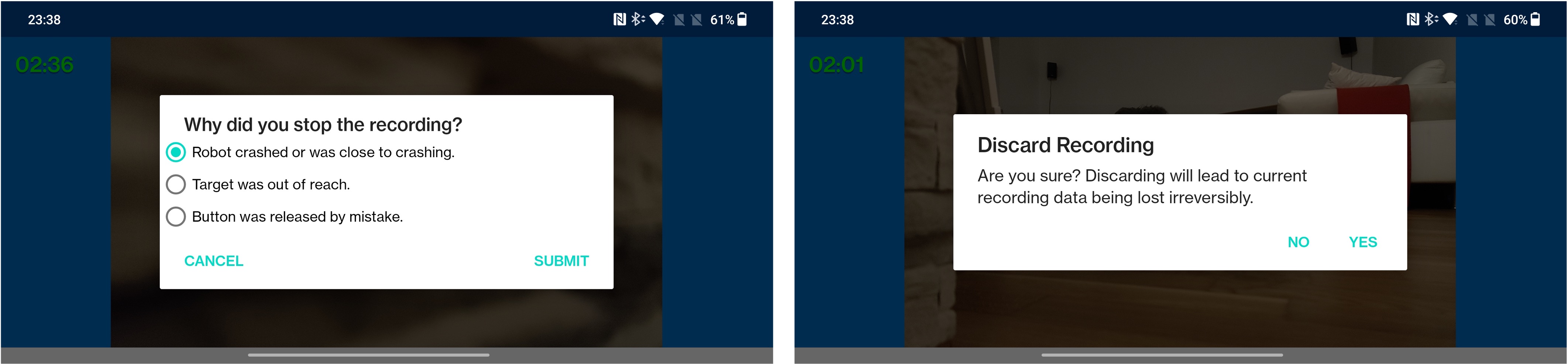}
    \caption{Screenshots of the smartphone app screens displayed if a recording is stopped by manual intervention.}
    \label{fig:screens_manual_stop}
\end{figure*}

A recording session is automatically stopped if the goal is reached or a collision is detected by the bump sensors; in that case we automatically assign a goal reward of $5$ or a collision penalty of $-1$. We also automatically detect several failure cases to ensure a high quality of recordings (see \Cref{fig:app_screens_failure}). We use the ARCore-based state estimation to ensure that the start and goal are at least 2 meters apart and that the robot is not picked up during the recording; if either is detected we cancel the recording and ask the user to start a new recording. If ARCore tracking is lost, for example due to bad lighting or fast motion, we also stop the recording. If it happens during automatic drive, we still allow the user to upload it and assign a penalty of $-0.5$. This encourages the policy to adapt to the limitations of ARCore. In addition to these automatic checks, we allow the user to stop the recording during automatic drive by releasing a button on the remote control. In this case, the user needs to select the stop reason and then upload the recording as usual. The user can also cancel and discard a recording at any time by pressing another button on the remote control. Both of these screens are show in \Cref{fig:screens_manual_stop}.

\subsubsection{Localization}
We utilize the ARCore API provided by Android to determine the relative position and orientation of the robot in the environment. ARCore uses a combination of visual feature tracking and data from the smartphone's inertial sensors to track the phone position. ARCore also has a depth API which we utilize to get the depth information for the latest RGB image obtained from the smartphone camera. 

\subsubsection{Data collection}
Before users can collect data, they need to complete the appropriate tutorial. After successful completion, the app checks for available recording requests from the cloud backend. If there are available requests, users can start a recording to collect data. The app can record synchronized data from sensors mounted on the robot, the smartphone's inertial sensors, cameras, and the state estimation and depth provided by ARCore. This can be configured according to requirements.
In our recording task, the user places the robot in the initial position and manually drives it to a feasible goal location within the environment using the shortest path while avoiding obstacles. %
If online learning is enabled, the current navigation policy then attempts to drive the robot back to the starting position while avoiding obstacles. Once the recording (one episode) is completed, the user can review a video of the recording and decide to upload it or to discard it. To guide the user through the data collection process and to provide alerts in failure cases, we have integrated on-screen dialogues and audio feedback.

\subsubsection{Navigation policies}
Our app supports two modes for the execution of navigation polices. In the first mode, users can select a policy served through the backend for evaluation. In the second mode, the policy is executed as part of the online learning process. In both cases, the user needs to actively hold a button on the remote control during the automatic drive of the robot for safety. Release of the button causes an immediate stop. The robot will also stop automatically if there is an error in the connection between phone and robot or remote control, a collision is detected or if ARCore stops tracking. 
During policy execution, observations are fed to the policy to produce controls for the robot. Observations include the current pose of the robot, the pose of the goal and either a RGB or depth image. The high-dimensional images are first fed into a pre-processor to produce a more efficient representation before being passed to the policy. For example, we use either MobileNetV2 \cite{howard2017mobilenets} or our perception backbone (see \Cref{sec:perception_backbone}) to process RGB images or extract rays from depth images. In the online learning setting, the policy inputs and outputs and the stop reason (for computing the reward) are uploaded to the cloud backend for consumption by the reinforcement learning framework that updates the policy.

\subsection{Remote Control}
\label{sec:remote}
The smartphone can be paired with a Bluetooth remote control to be used for navigating through the app and to manually drive the robot. This enables data collection of labeled image-control pairs during manual control by a user.

\section{Cloud Backend}
\label{sec:backend}

The backend acts as the interface between the robot fleet and learning framework by facilitating data collection and serving navigation policies (\Cref{fig:system_overview}-B).
We realize our backend with Firebase (\url{firebase.google.com}) and utilize the following Firebase modules.

\subsection{Authentication}
We require authentication by the workers via username and password. This prevents a misuse of the app and allows an enforcement of certain security rules, \eg such that access to recordings is limited to their creators.

\subsection{Cloud Firestore}
Cloud Firestore is a scalable NoSQL cloud database that provides synchronization via real-time listeners. We configure a Firestore database to upload recording requests from the learning framework (see \Cref{sec:rl_framework}) and to store corresponding recordings from the workers (see \Cref{sec:fleet}). Here, a recording request contains a link to the policy file, a description of the recording task, and a list of permitted users. On upload, all permitted users are notified about a new request in real-time. When a user starts working on a request, the request is blocked for other users until the recording is finished or a timeout is reached.
On successful upload of a recording, we inform the learning framework via a real-time listener. The learning process consumes the recording and might create new requests as needed; see \Cref{sec:rl_framework}.

\subsection{Cloud Storage}
We use Cloud Storage to store policies (TensorFlow Lite models) and the collected recordings.

\subsection{Cloud Functions}
We apply Cloud Functions to delete a recording request on upload of a corresponding recording file. Moreover, we implement functions to collect statistics about the number and length of recordings per worker. %

\section{Reinforcement Learning Framework}
\label{sec:rl_framework}

Our reinforcement learning framework is designed for online learning with a fleet of robots deployed in the real world. This requires running policies on the smartphones. To this end, we implement our learning framework with TF-Agents~\cite{TFAgents}, a TensorFlow-based library for reinforcement learning. In TF-Agents, all policies are implemented in TensorFlow which allows a conversion to TensorFlow Lite models. This conversion is key to run those policies on mobile devices such as smartphones.

The basic architecture of TF-Agents is shown \Cref{fig:system_overview}-C. The \emph{driver} is responsible for executing a policy called the \emph{collect policy}. The output of such policy runs are trajectories that are fed into the \emph{replay buffer}. Each \emph{trajectory} contains a tuple of observations, actions, rewards, and additional meta-data, \eg, if a step caused the end of an episode. 
We implement a subclass of the driver class (\emph{FirebaseDriver}) that allows to publish policies to the Firebase backend and to collect the resulting trajectories; refer to \Cref{sec:backend}. It also handles the conversion of the collect policy to a TensorFlow Lite model and the parsing of the recorded log file to TF-Agents trajectories. 

Each log file contains a full episode executed on a worker's smartphone and robot. For each step of this episode, we record the observation, the action taken by the collect policy, and additional information such as the relative position of the robot to the target (based on ARCore measurements). This additional information allows us to assign rewards in a post-processing step in the learning framework. In this way, we have greater flexibility in shaping the reward function compared to computing the reward directly on the smartphone. (We do not require to ship a new app version whenever the reward function changes.)

\section{Imitation Learning Framework}
\subsection{Perception backbone}
\label{sec:perception_backbone}
Given an RGB image, the objective of the perception backbone is to segment the floor. There are four parts to train the backbone:

\subsubsection{Data}
We use RGB images from the real world collected by the robot fleet and generate approximate ground-truth segmentation masks using our automatic labelling pipeline. In addition, we also use a dataset with 12k RGB images and GT segmentation from SPEAR. We use a simple U-Net to train our perception backbone.

\subsubsection{Automatic labeling pipeline}
\label{sec:auto_labeling}
\begin{figure*}[!htb]
    \centering
    \includegraphics[width=\textwidth]{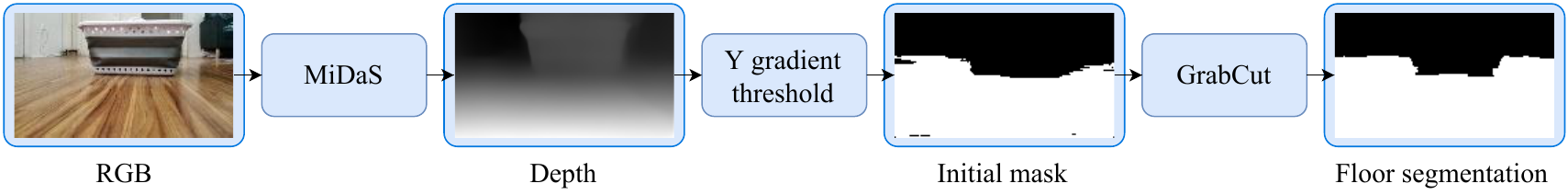}
    \caption{Automatic labeling pipeline to generate the ground truth for floor segmentation.}
    \label{fig:autolabeling}
\end{figure*}
\Cref{fig:autolabeling} shows the automatic labeling pipeline. First, an RGB image is passed through a MiDaS monocular depth estimation model \cite{ranftl2020towards} which outputs pixel-wise relative depth; we use MiDaS 3.1 with the LeViT \cite{graham2021levit} backbone. Since the floor is a horizontal plane in the environment we can get a reasonable segmentation with an appropriate threshold on the gradient of depth image in the y-direction. We use this approach to generate an initial binary mask. This binary mask is used to initialize the foreground extraction algorithm GrabCut \cite{rother2004grabcut}. Using the original RGB image and the initial mask, Grabcut generates the final floor segmentation mask which we use as the ground truth to train our perception backbone.

\begin{figure}[!htb]
    \centering    \includegraphics[width=\columnwidth]{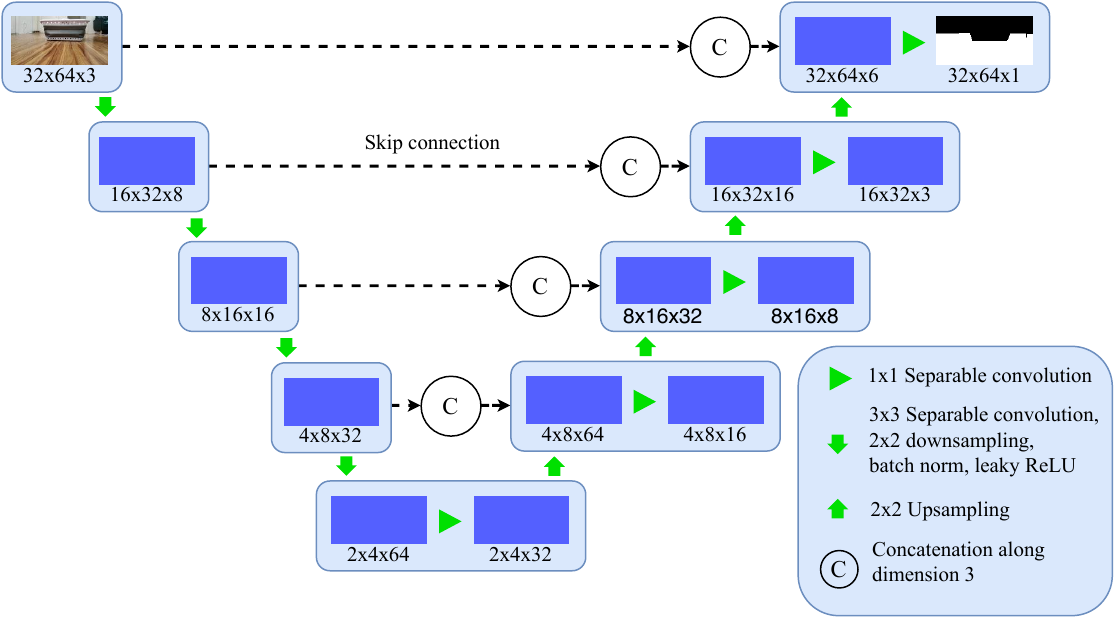}
    \caption{U-Net architecture for our floor segmenation pre-processor.}
    \label{fig:unet}
\end{figure}

\subsubsection{Network}
\label{sec:unet_model}
We use a light-weight U-Net architecture \cite{ronneberger2015u} as shown in \Cref{fig:unet}. Efficiency is key for real-time inference on the robots since the smartphone capabilities may vary among crowd workers. The RGB images are aquired at a resolution of 90x160 and then scaled down to 32x64 before be passed to the U-Net.

\subsubsection{Training}
\label{sec:unet_training}
We use the diceloss $\large(1-\frac{2 \times \text{intersection}}{\text{total area}}\large)$ and Adam Optimizer \cite{kingma2014adam} to train the U-Net model for 50 epochs and keep the best checkpoint based on the validation loss. 

\subsection{Conditional Imitation Learning}
We implement conditional imitation learning \cite{codevilla2018end} with a PilotNet \cite{bojarski2016end} backbone for efficient inference on smartphones. We inject a relative vector from the robot to the goal location as the condition by concatenating it with the second layer of the MLP in PilotNet. We validate in simulation that this architecture works well with expert demonstrations. In this paper, we use it as a behaviour cloning baseline for end-to-end regression of controls based on data collected by the crowd workers. %

\section{Simulation Platforms}
\label{sec:simulation}

\begin{figure*}[!htb]
            \centering
    \begin{subfigure}{0.55\linewidth}
        \begin{minipage}[b]{.49\columnwidth}
            \centering
                \includegraphics[width=\columnwidth]{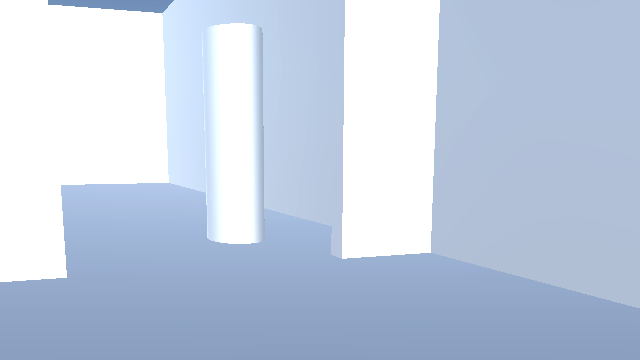} \\ 
                \vspace{3pt}
                \includegraphics[width=\columnwidth]{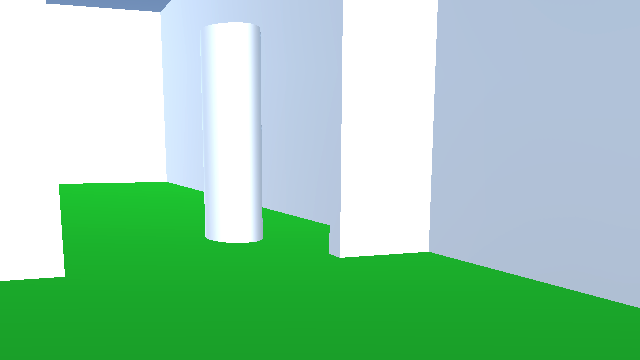} 
                \vfill
        \end{minipage}
        \hfill
        \begin{minipage}[b]{.49\columnwidth}
            \centering
                \includegraphics[width=\columnwidth]{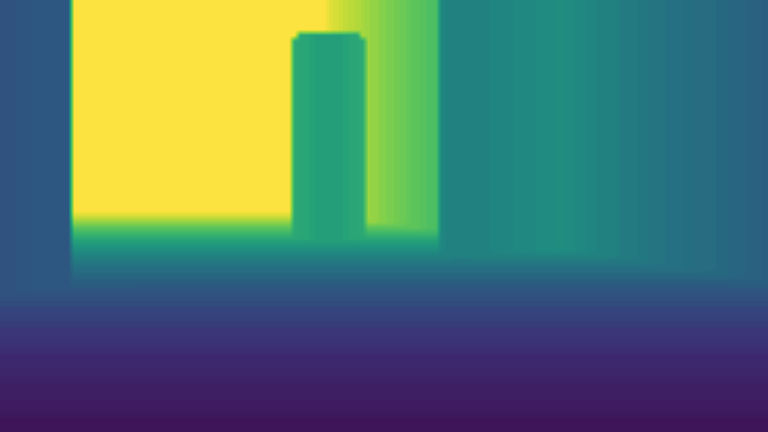} \\ 
                \vspace{3pt}
                \includegraphics[width=\columnwidth]{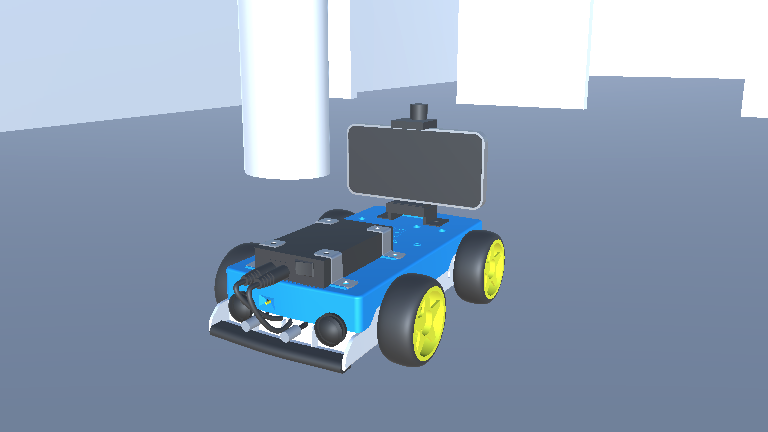}
                \vfill
        \end{minipage}
        \caption{Egocentric RGB, depth and segmentation observations, and a 3rd person view of the OpenBot agent.}
        \label{fig:unity_env_and_depth}
    \end{subfigure}
     \hfill
    \begin{subfigure}{0.43\linewidth}
        \includegraphics[clip, trim=0cm 0.15cm 0cm 1.8cm, width=\columnwidth]{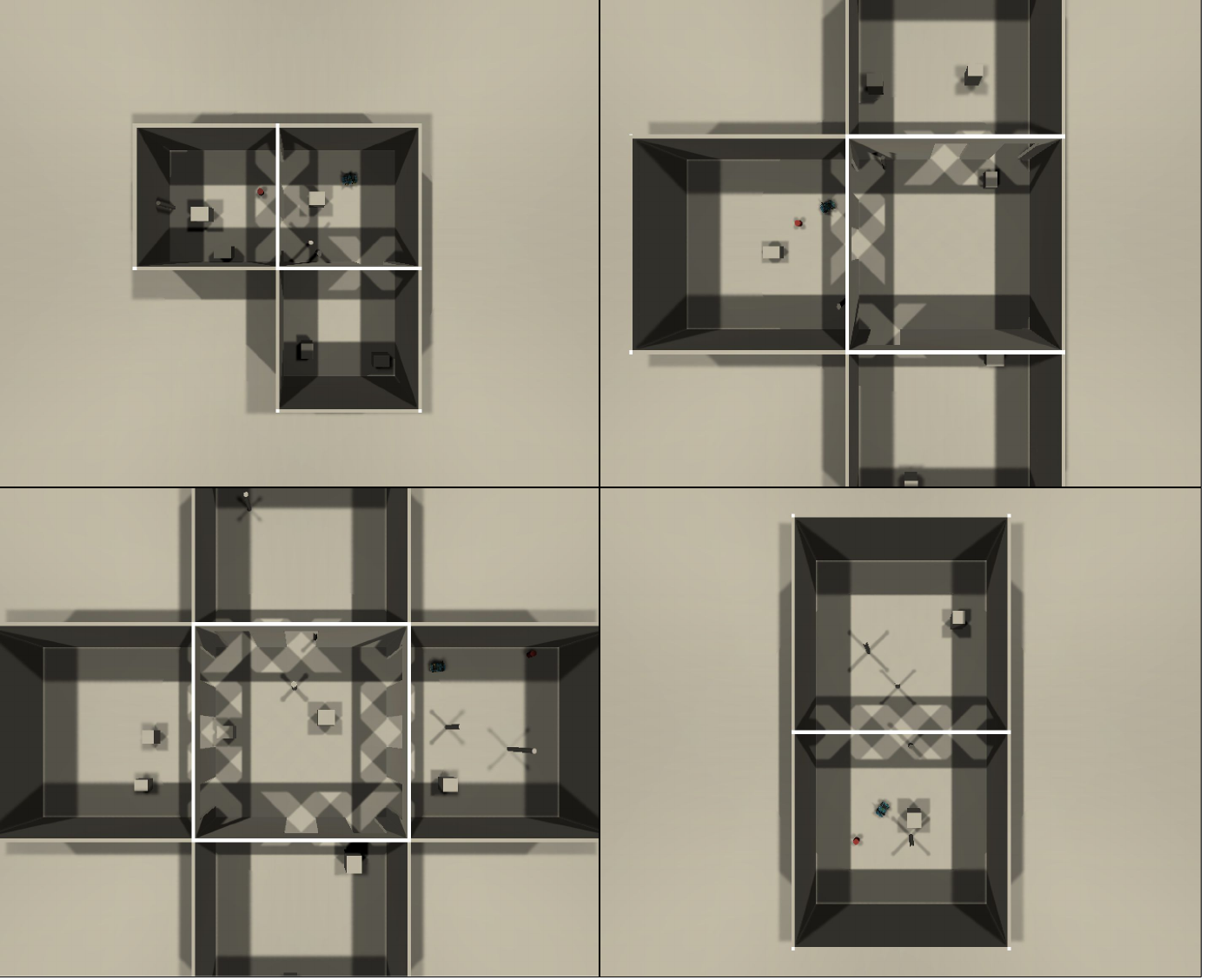}
        \caption{Procedurally generated layouts with random room size, obstacle shape, goal location and robot position.
        }
        \label{fig:unity_procedural_envs}
    \end{subfigure}
    \caption{Basic simulation based on Unity ML-Agents for pretraining of RL policies.}
    \label{fig:unitysim}
\end{figure*}

\subsection{ProceduralSim}

We have built ProceduralSim on top of Unity ML-Agents~\cite{juliani2018unity}. It contains a 3D geometric model of our robot and appropriate control interfaces. The configuration-space control interface accepts a 2D input $(\tau_l, \tau_r)$, where $\tau_{l, r} \in \left[-1, 1\right]$, to set the left and right wheel torques. On the other hand, the task-space control interface accepts a 3D input $(x, y, \theta)$, where $x,y \in \left[-1, 1\right]$ are (scaled) relative position targets, and $\theta$ is the relative heading target. This input is then internally converted to wheel torques using the unicycle model of differential drive robots~\cite{Lynch2017ModernRM, unicycle_model}. The simulation platform also supports adding a virtual camera with configurable intrinsics to the robot that can observe RGB, depth, and floor segmentation images, as shown in \Cref{fig:unity_env_and_depth}.

In addition, ProceduralSim also includes procedural environment generation. The environment can be generated either by specifying ranges within which environment parameters like number of rooms and connecting doors, type and height of obstacles (cubes and cylinders) and start and goal position of the robot, will be randomly sampled, or by directly specifying the environment parameters for deterministic evaluation. Examples of generated environments are shown in Figure~\ref{fig:unity_procedural_envs}.

ProceduralSim has a Gym-style class~\cite{openai_gym} for use with RL training frameworks. This class instantiates a new environment when reset and accepts actions in one of the two formats described above. It returns an observation vector consisting of the distance to the goal, the angle difference between robot heading and the vector from robot to target, the flattened RGB, depth or segmentation image, and optionally, robot telemetry information like its global pose and collision status. This enables the use of ProceduralSim with standard RL training frameworks for easy experimentation and comparison with other methods.

\begin{figure*}[!htb]
    \begin{subfigure}{0.64\linewidth}
        \begin{minipage}[b]{.49\columnwidth}
            \centering
                \includegraphics[width=\columnwidth]{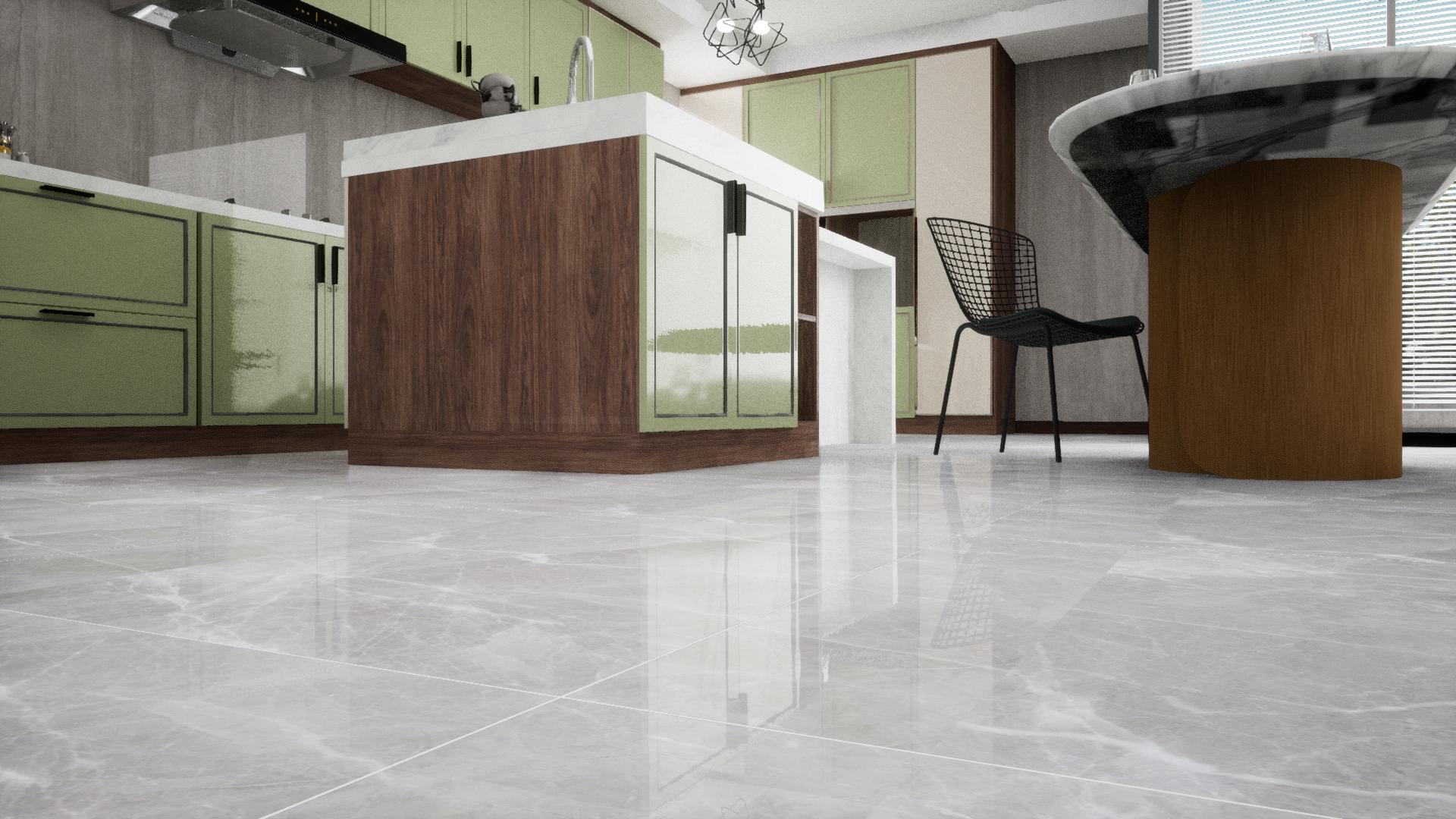} \\
                \vspace{3pt}
                \includegraphics[width=\columnwidth]{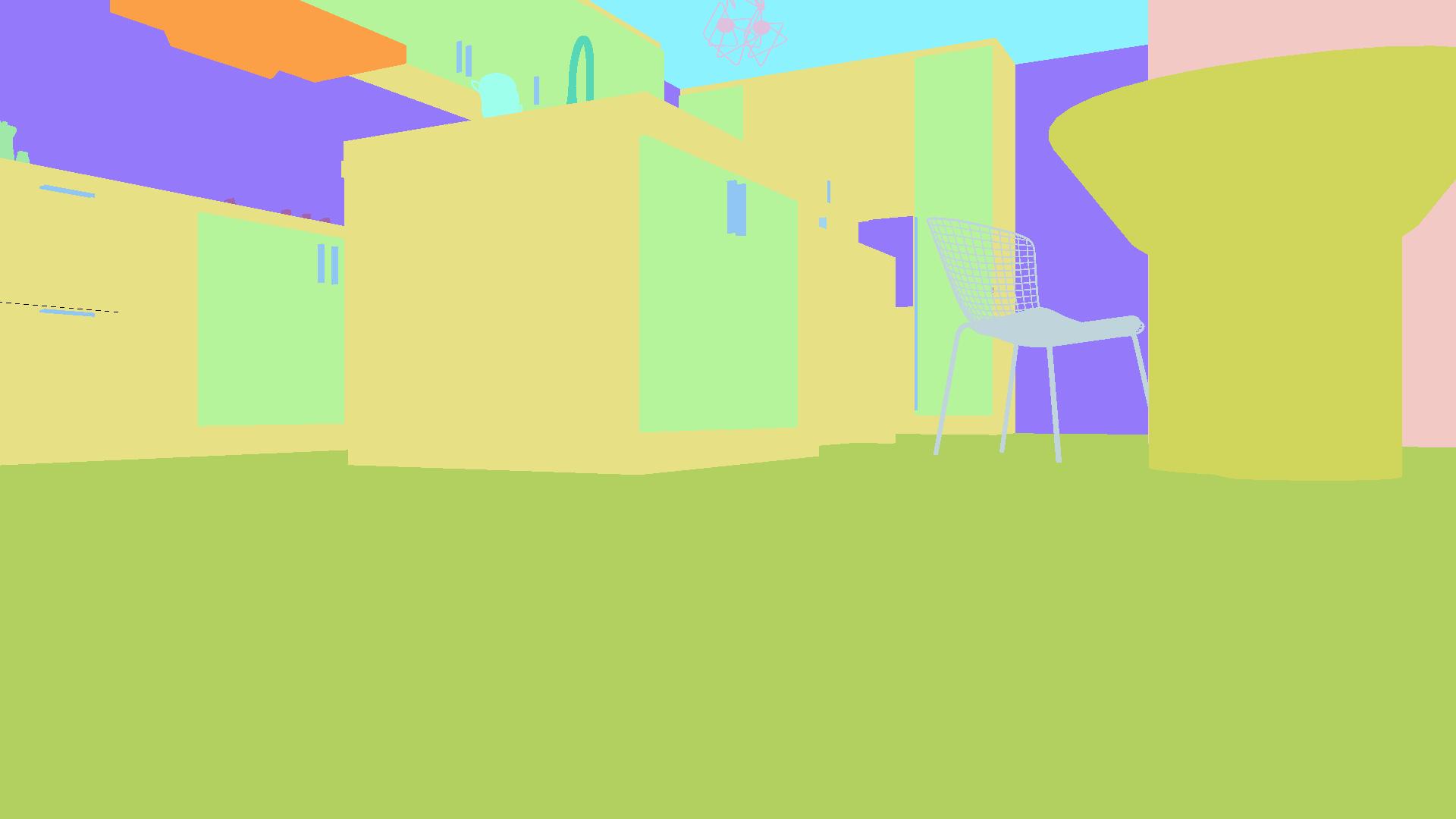} 
        \end{minipage}
        \hfill
        \begin{minipage}[b]{.49\columnwidth}
            \centering
                \includegraphics[width=\columnwidth]{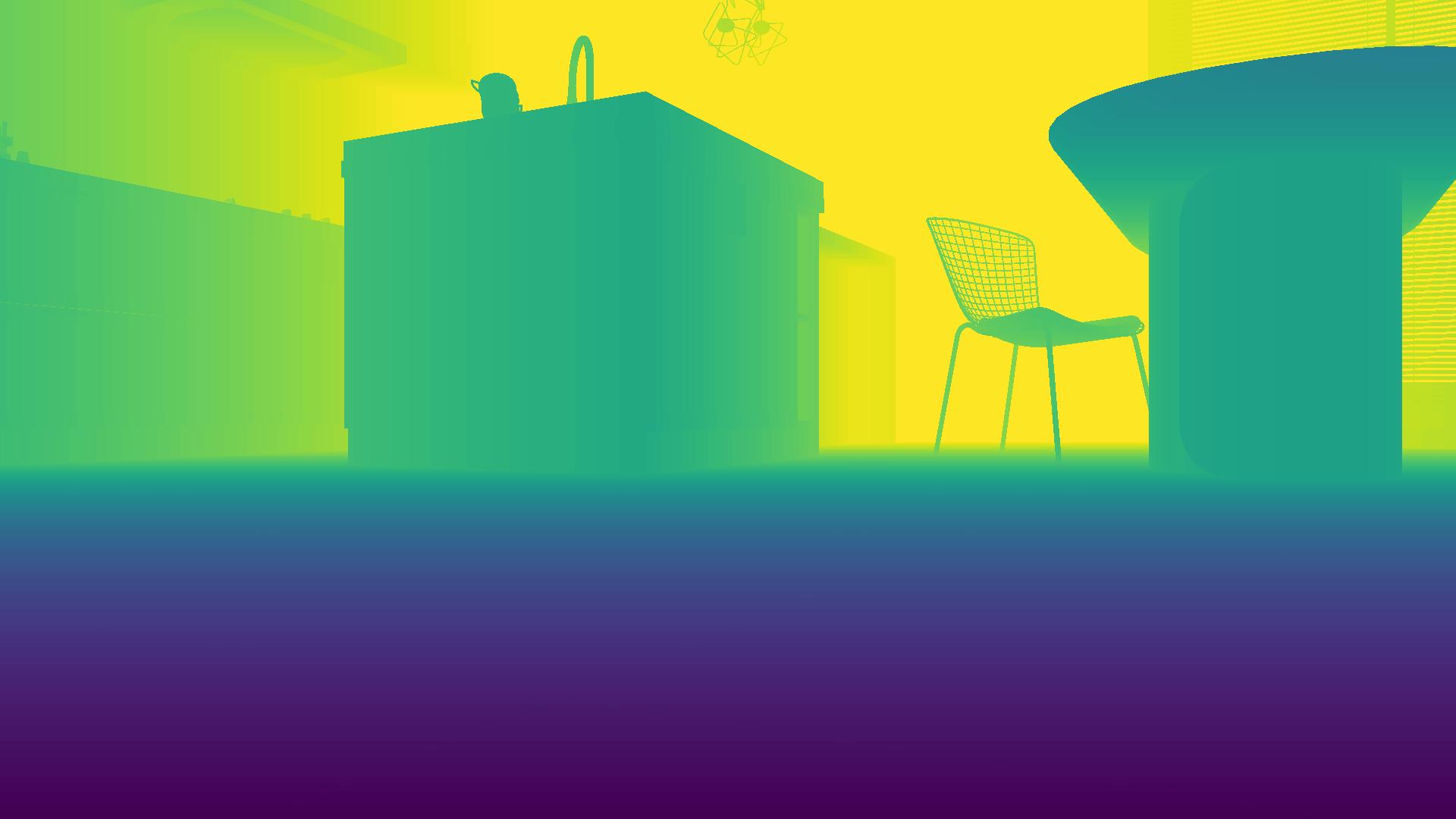} \\
                \vspace{3pt}
                \includegraphics[width=\columnwidth]{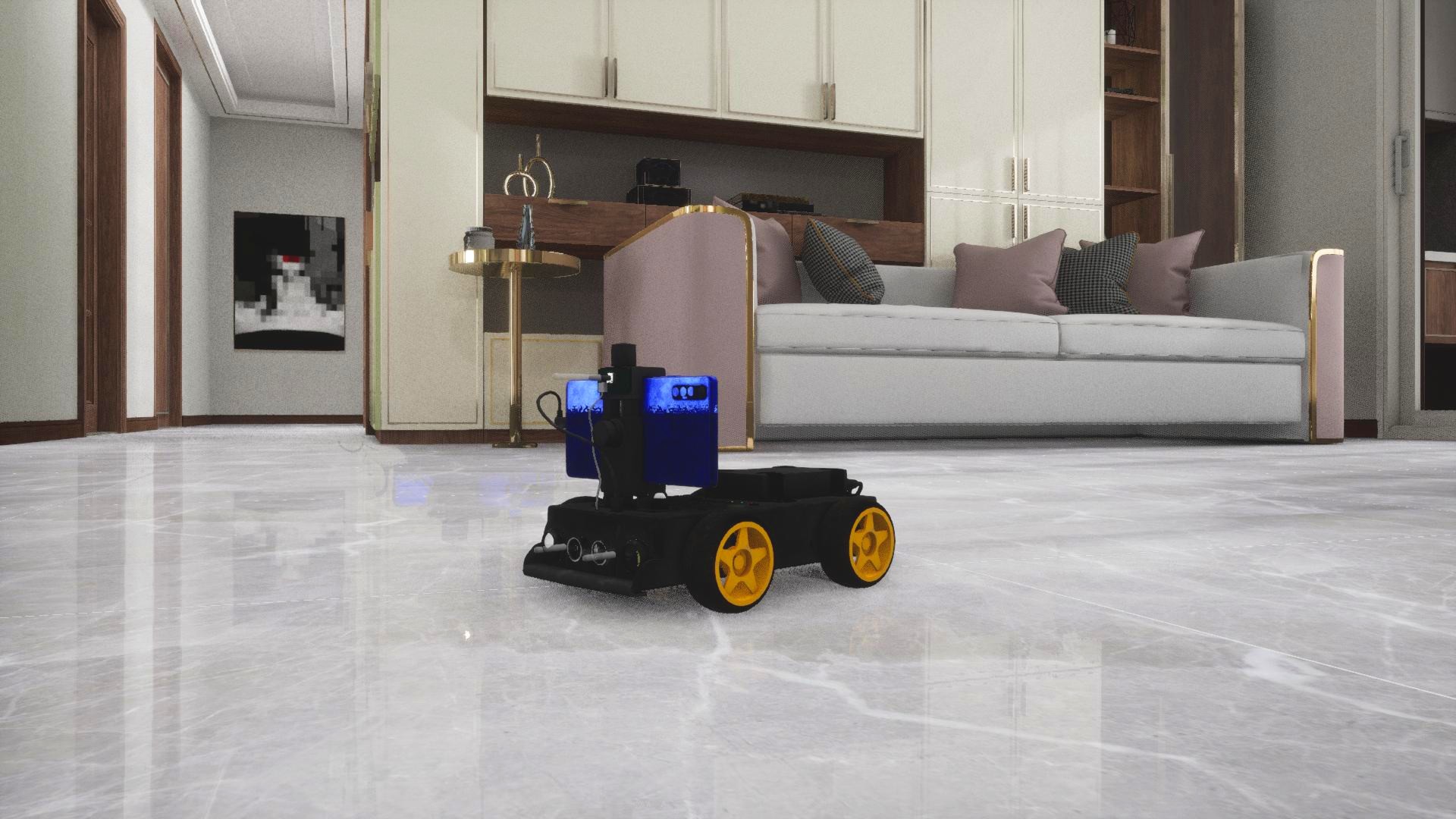}
        \end{minipage}
        \caption{Egocentric RGB, depth and segmentation observations, and a 3rd-person view of the simulated OpenBot agent.}
        \label{fig:spear_env_and_depth}
    \end{subfigure}
    \hfill
    \begin{subfigure}{0.34\linewidth}
        \includegraphics[width=\columnwidth]{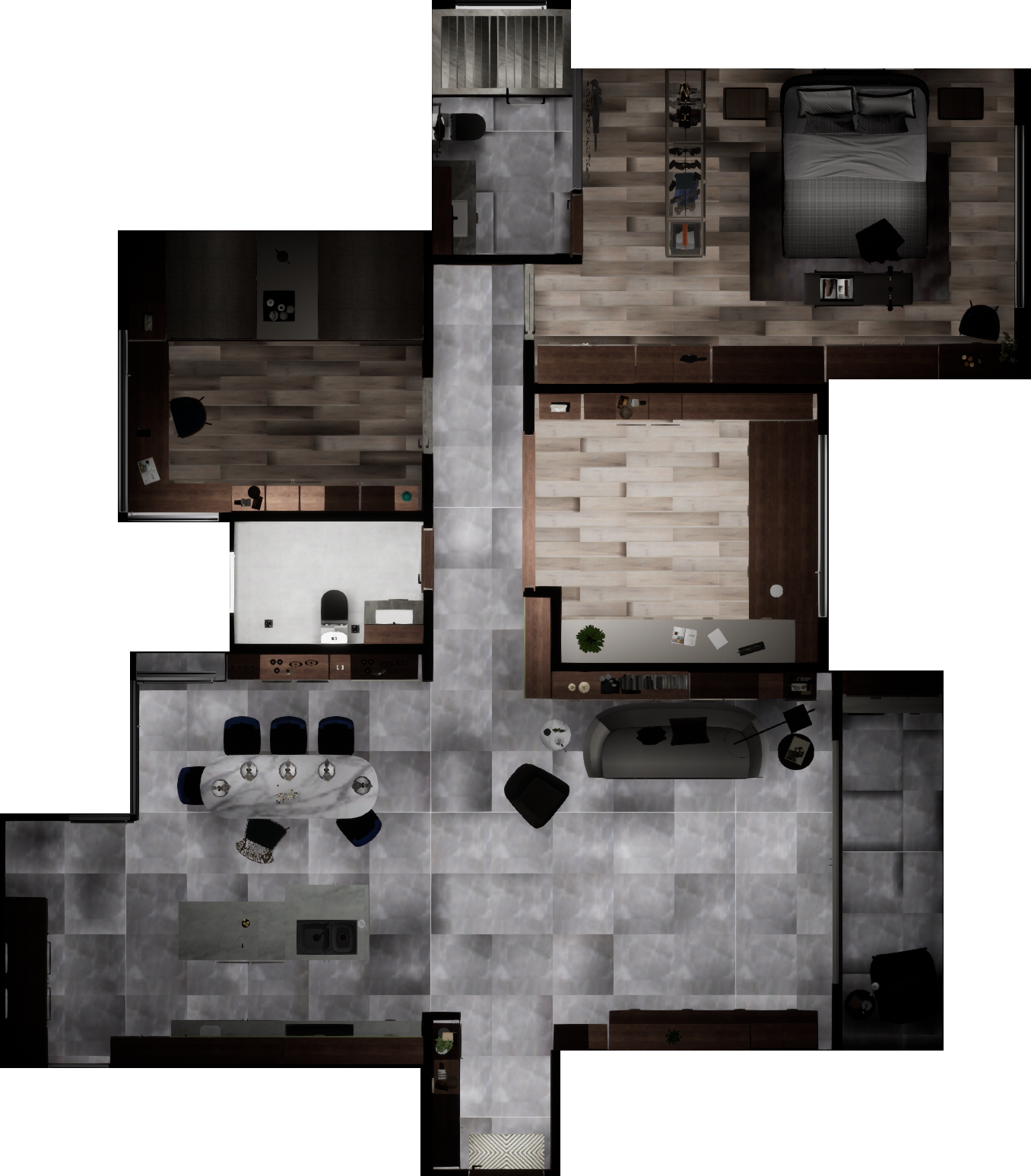}
        \caption{Top view of the SPEAR environment.}
        \label{fig:spear_realistic_env}
    \end{subfigure}
    \caption{Photo-realistic simulation based on SPEAR for evaluation and fine-tuning of RL policies.}
    \label{fig:spearsim}
\end{figure*}

\subsection{SpearSim}

SPEAR (Simulator for Photo-realistic Embodied AI Research) \cite{spear} is a recent simulation platform for the development and training of embodied AI systems. It features a range of virtual sensors that simulated agents can be equipped with. Beyond RGB images, the camera sensor can also be set up to provide groundtruth semantic segmentation and depth images. SpearSim also provides photo-realistic indoor environments and an OpenAI Gym interface, a standard API for reinforcement learning. We use it in our system for evaluation of the complete navigation stack, consisting of our perception backbone and a control policy, before large-scale deployment on real robots.

\end{appendices}
\end{document}